\documentclass[10pt]{article} %
\usepackage[preprint]{tmlr}
\usepackage{amsmath,amsfonts,bm}

\usepackage[utf8]{inputenc} %
\usepackage[T1]{fontenc}    %
\usepackage{hyperref}       %
\definecolor{darkgray}{RGB}{60,60,60}
\hypersetup{
    colorlinks=true,
    linkcolor=darkgray,
    urlcolor=darkgray,
    citecolor=darkgray
}
\usepackage{url}            %
\usepackage{booktabs}       %
\usepackage{amsfonts}       %
\usepackage{nicefrac}       %
\usepackage{microtype}      %
\usepackage{xcolor}         %
\usepackage{graphicx}
\usepackage{wrapfig}
\usepackage{multirow}

\usepackage{algorithm}
\usepackage{algpseudocode}

\newcommand{\ours}{{\textsc{STeP}}}

\usepackage{xcolor}
\definecolor{rebuttalcolor}{HTML}{0047AB}
\newcommand{\rebuttal}[1]{{\color{rebuttalcolor}#1}}
\newcommand{\rebuttalon}{\color{rebuttalcolor}}   %
\newcommand{\kine}{\texttt{kine}}
\newcommand{\suppchange}[1]{#1}
\newcommand{\suppchangeon}{}                       %

\usepackage{amsmath,amsfonts,bm,amssymb,mathtools,amsthm}
\usepackage{color,xcolor,placeins}
\usepackage{thm-restate}

\usepackage{hyperref}       %
\hypersetup{
  colorlinks,
  linkcolor={blue!50!black},
  citecolor={blue!50!black},
  urlcolor={blue!80!black}
}
\definecolor{orange}{HTML}{ff6c0c}
\definecolor{blue}{HTML}{1f77b4}
\definecolor{Gray}{gray}{0.85}
\definecolor{LightCyan}{rgb}{0.88,1,1}

\usepackage{url}
\usepackage{xkcdcolors}
\usepackage{mdframed}

\usepackage{microtype}
\usepackage{graphicx}
\usepackage{subcaption}
\usepackage{latexsym}
\usepackage{sidecap}
\usepackage{caption}
\usepackage{booktabs} %
\usepackage{amsfonts}       %
\usepackage{nicefrac}       %
\usepackage{comment}
\usepackage{enumitem}
\usepackage{wrapfig,tikz}
\usepackage{longtable}
\usepackage{bigstrut, tabularx, multirow, makecell, diagbox}
\usepackage[export]{adjustbox}
\usepackage{float}
\usepackage{stackrel}
\usepackage{color}
\usepackage{colortbl}
\usepackage{theoremref}
\usepackage{cases}
\usepackage{stmaryrd}
\usepackage{mathabx}
\usepackage{dsfont}
\usepackage{arydshln}
\usepackage{algorithm}
\usepackage{algpseudocode}

\makeatletter
\usepackage{xspace}
\def\@onedot{\ifx\@let@token.\else.\null\fi\xspace}
\DeclareRobustCommand\onedot{\futurelet\@let@token\@onedot}

\definecolor{blue1}{RGB}{0,128,255}
\definecolor{blue3}{RGB}{0,0,128}
\definecolor{darkpastelgreen}{rgb}{0.01, 0.75, 0.24}
\definecolor{cerulean}{rgb}{0.0, 0.48, 0.65}

\def\vs{\emph{vs}\onedot}

\definecolor{darkgreen}{rgb}{0,0.6,0}

\def\Figref#1{Figure~\ref{#1}}
\def\Tabref#1{Table~\ref{#1}}

\def\Secref#1{Sec.~\ref{#1}}
\def\eqref#1{equation~(\ref{#1})}
\def\Eqref#1{Equation~(\ref{#1})}
\def\Algref#1{Algorithm~\ref{#1}}

\def\1{\bm{1}}

\def\rvtheta{{\bm{\theta}}}

\def\rvm{{\mathbf{m}}}
\def\rvn{{\mathbf{n}}}

\def\rvp{{\mathbf{p}}}

\def\rvs{{\mathbf{s}}}

\def\rvx{{\mathbf{x}}}

\def\rvy{{\mathbf{y}}}
\def\rvz{{\mathbf{z}}}
\def\rvzhatzero{{\hat{\mathbf{z}}_0}}

\def\rvI{{\mathbf{I}}}
\def\rvV{{\mathbf{V}}}

\def\vzero{{\bm{0}}}

\def\vtheta{{\bm{\theta}}}

\def\vs{{\bm{s}}}

\def\mI{{\bm{I}}}

\DeclareMathAlphabet{\mathsfit}{\encodingdefault}{\sfdefault}{m}{sl}
\SetMathAlphabet{\mathsfit}{bold}{\encodingdefault}{\sfdefault}{bx}{n}

\newcommand{\R}{\mathbb{R}}

\let\rebuttal\steprevision
\renewcommand{\rebuttalon}{}

\title{\textsc{STeP}: Solving Scientific Video Inverse Problems\\ with Spatiotemporal Diffusion Priors}

\author{
  \name Bingliang Zhang\textsuperscript{*} \\
  \addr California Institute of Technology
  \AND
  \name Zihui Wu\textsuperscript{*} \\
  \addr California Institute of Technology
  \AND
  \name Berthy T. Feng \\
  \addr California Institute of Technology
  \AND
  \name Yang Song \\
  \addr Meta
  \AND
  \name Yisong Yue \\
  \addr California Institute of Technology
  \AND
  \name Katherine L. Bouman \\
  \addr California Institute of Technology \\
  \small\textsuperscript{*}Equal contribution.
}

\begin{document}

\maketitle

\begingroup
\renewcommand{\thefootnote}{}
\begin{NoHyper}
\footnotetext{We used GPT to improve the grammar and wording of the manuscript.}
\end{NoHyper}
\endgroup

\begin{abstract}
\rebuttal{In scientific video inverse problems, such as black hole video reconstruction and dynamic MRI, a video of a dynamic phenomenon must be recovered from sparse, indirect measurements. These measurements alone often cannot determine how the scene evolves over time. Diffusion-based video solvers that apply an image diffusion prior to each frame and couple the frames through heuristics, such as shared or flow-warped noise, struggle when the measurements carry little information about the temporal evolution. We present \ours, a plug-and-play approach that replaces these heuristics with a spatiotemporal diffusion prior learned from data. \ours~builds the prior with standard video diffusion techniques, extending a latent image diffusion model with temporal layers and fine-tuning it jointly on images and a limited number of videos, and pairs it with an existing posterior sampler. On black hole video reconstruction from simulated Event Horizon Telescope observations, \ours~improves PSNR by 2.62~dB over the strongest image diffusion baseline, recovers temporal dynamics that the baselines miss, and produces distinct reconstructions that fit the data comparably well. Joint training with images helps at every training-set size we test. On dynamic MRI, the same recipe improves PSNR by 3.83~dB over the strongest baseline at 8$\times$ acceleration.}

\end{abstract}    
\section{Introduction}
\label{sec:intro}

Reconstructing high-quality videos from time-varying measurements is a core challenge in many scientific applications, such as black hole video reconstruction~\citep{akiyama2022firstIII} and dynamic magnetic resonance imaging (MRI)~\citep{gamper2008compressed}, where recovered spatiotemporal features directly influence downstream scientific analysis or medical interpretation.
Unlike standard imaging tasks, these problems involve recovering a high-dimensional spatiotemporal signal from measurements that are often sparse, indirect, and corrupted by noise, making the inference task fundamentally ill-posed. In many cases, substantial spatial and temporal information is irreversibly lost during data acquisition, requiring strong prior knowledge to constrain the solution space.

Diffusion models (DMs)~\citep{ho2020denoising, song2021scorebased} have recently emerged as state-of-the-art tools for modeling complex image and video distributions, providing powerful learned priors for solving inverse problems~\citep{chung2023diffusion, zheng2025inversebench}.
Motivated by their success, a growing body of work has explored diffusion-based methods for video inverse problems. However, most existing approaches are designed for video editing or photographic restoration tasks~\citep{daras2024warped, kwon2025solving, kwon2024visionxlhighdefinitionvideo, yeh2024diffir2vrzerozeroshotvideorestoration}, where the underlying spatiotemporal structure is either given, externally specified, or largely preserved in the measurements. In these settings, temporal dynamics can often be inferred directly from the measurements or auxiliary cues.

These methods typically rely on image diffusion models (IDMs) that process frames independently, while enforcing temporal consistency through heuristic coupling mechanisms such as correlated noise injection or optical-flow-based warping~\citep{kwon2025solving, chang2024how, daras2024warped}.
These strategies are effective for restoration problems where measurements retain rich spatiotemporal information, but we find that they struggle in scientific video inverse problems, where measurements may contain limited spatial structure and provide little direct information about temporal evolution.

Scientific video inverse problems differ fundamentally from conventional video restoration.
In these problems, the measurement process, which is often dictated by complex physical systems, can hide or distort both spatial and temporal information. As a result, temporal dynamics cannot be directly recovered from the measurements, and multiple videos with very different motion patterns may explain the same observed data. \rebuttal{This makes heuristic temporal coupling unreliable and calls for priors that explicitly model joint spatiotemporal structure. It also shifts the goal from finding a single visually plausible reconstruction to exploring multiple videos that are consistent with the data.}

Plug-and-play diffusion prior (PnPDP) methods offer a general and flexible framework for inverse problems by decoupling the learned prior from the data fidelity term.
They have demonstrated strong performance across a wide range of applications, including natural image restoration~\citep{wang2022zero, song2021scorebased, kawar2022denoising, chung2023diffusion, lugmayr2022repaint, zhu2023denoising}, medical imaging~\citep{jalal2021robust, song2022solving, chung2022score, chung2022mr, zhang2024improving}, and physics-based inverse problems~\citep{zheng2024ensemblekalmandiffusionguidance, feng2023score, sun2024provable, wu2024principled}.
Despite being compatible with videos in principle, existing PnPDP methods have been predominantly studied in static image settings. \rebuttal{Learned video priors, both spatiotemporal denoisers in plug-and-play schemes~\citep{monod2022video} and video diffusion models~\citep{kwon2024seeing, spagnoletti2025lvtino, kwon2026avis}, have mostly been applied to natural videos, for which training data are abundant (Sec.~\ref{sec:background}). In scientific domains, training videos are typically scarce (e.g., several hundred simulations for black holes), and it is not established how well video diffusion priors learned from such data perform on highly ill-posed scientific video inverse problems.}

\begin{figure}
    \centering
    \includegraphics[width=\textwidth]{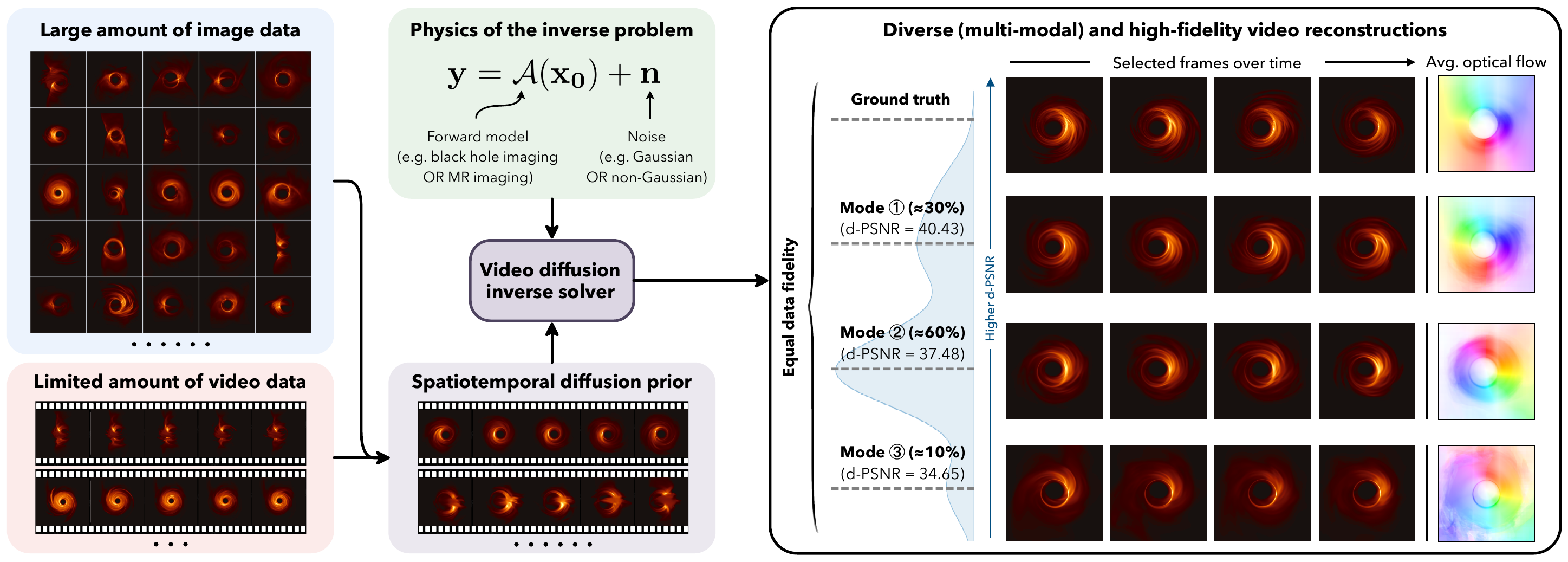}
    \caption{\textbf{Overview of {\ours}.} \underline{Left:} \ours~combines the physical model of the target problem with a spatiotemporal diffusion prior that directly characterizes the video distribution. \rebuttal{The prior is obtained by training an image diffusion model, extending it with temporal layers, and fine-tuning it jointly on images and a limited number of videos. \underline{Right:} \ours~can generate diverse solutions for a single black hole observation that fit the measurements comparably well (data misfit from 1.01 to 1.14, versus mean misfits of 1.38 to 1.74 for the IDM-based baselines in Tab.~\ref{tab:main-table}). Percentages give the fraction of 100 posterior samples assigned to each mode, and d-PSNR (Sec.~\ref{sec:metrics}) measures temporal agreement with the ground truth.}}
    \vspace{-15pt}
     \label{fig:teaser}
\end{figure}

\rebuttal{In this paper, we present {\ours}, which plugs a learned spatiotemporal diffusion prior into a PnPDP sampler for scientific video inverse problems, so that temporal structure is learned from data rather than imposed through heuristics.
{\ours} builds on established components (Tab.~\ref{tab:prior-sources}): following standard practice in video generation~\citep{ho2022videodiffusionmodels, singer2023makeavideo, blattmann2023alignlatentshighresolutionvideo, wang2023modelscopetexttovideotechnicalreport}, we extend a latent image diffusion model with temporal layers and fine-tune it jointly on images and videos, and we sample the posterior with Decoupled Annealing Posterior Sampling (DAPS)~\citep{zhang2024improving}.
We study whether this combination is effective in a regime that diffusion-based video solvers have not addressed, with extremely sparse, nonlinear measurements and only several hundred training videos, and evaluate it in depth on black hole video reconstruction.
Our main contributions are:
\begin{itemize}[leftmargin=*,itemsep=0pt,topsep=2pt]
    \item We show that a spatiotemporal diffusion prior fine-tuned on several hundred simulated videos substantially improves black hole video reconstruction. On simulated Event Horizon Telescope (EHT) observations of Sagittarius A$^\ast$, {\ours} improves PSNR by 2.62~dB over the strongest image diffusion baseline, recovers temporal dynamics that these baselines miss even when the noise-warping baselines are given oracle optical flow (Fig.~\ref{fig:blackhole-imaging}), and produces distinct reconstructions that fit the data comparably well (Fig.~\ref{fig:teaser}).
    \item We find that joint training with images helps at every training-set size we test (\Secref{sec:effectiveness}).
    \item We apply the same recipe to dynamic MRI, whose spatiotemporal characteristics differ markedly from those of black hole videos, and improve PSNR by 3.83~dB over the strongest baseline at 8$\times$ acceleration.
\end{itemize}}

\begin{figure}[t!]
    \centering
    \includegraphics[width=\linewidth]{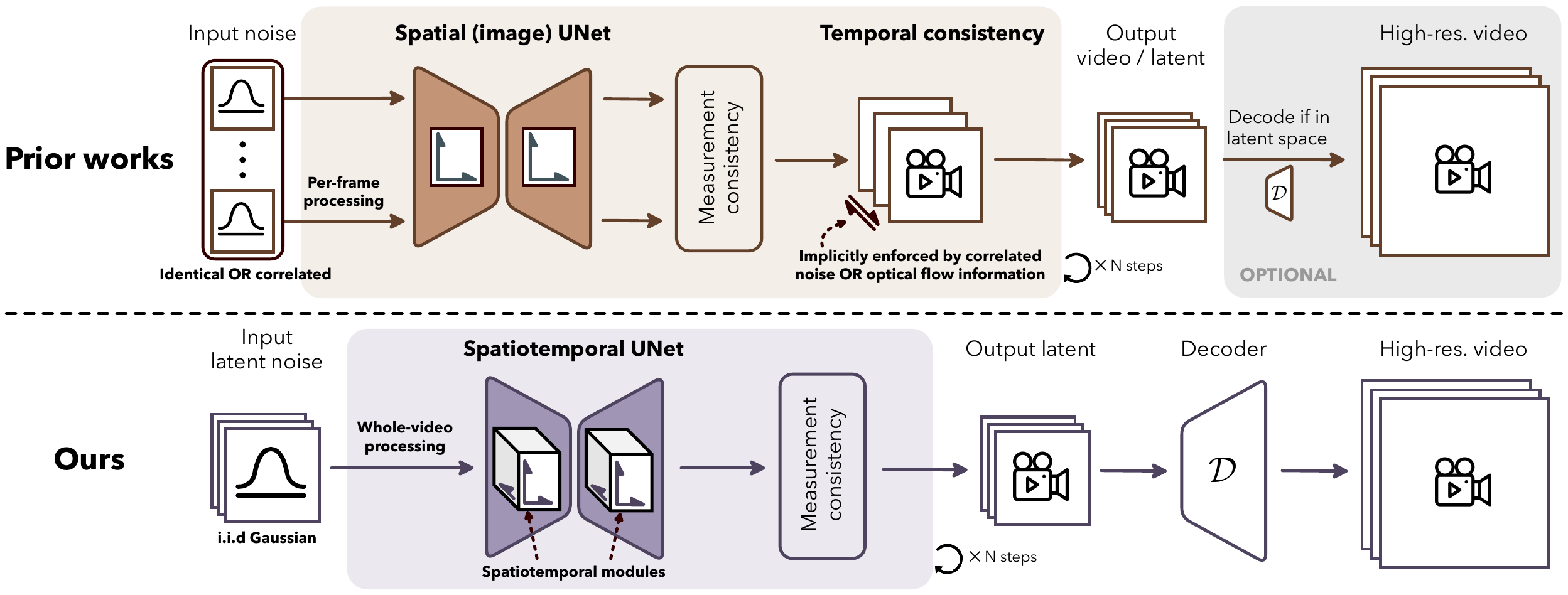}
    \caption{\textbf{A schematic comparison between existing IDM-based methods (\underline{top}) and \ours~(\underline{bottom}) for video inverse problems.} Bold text marks the key differences between them. While existing methods use an image diffusion model and enforce temporal consistency using simple heuristics or warping noise with optical flow, we directly learn a spatiotemporal diffusion prior. 
    }
    \vspace{-0.1in}
    \label{fig:method}
\end{figure}

\section{Background}
\label{sec:background}

\paragraph{Diffusion-based video inverse problems}
Diffusion-based approaches to video inverse problems have primarily focused on restoration and editing problems, such as super-resolution and JPEG artifact removal, for natural videos~\citep{chang2024how, daras2024warped, kwon2024visionxlhighdefinitionvideo, deng2025infiniteresolution, zou2025flair, kwon2025solving}.
\rebuttal{Many of these methods rely on an IDM to process each video frame and propose various techniques to enforce temporal consistency.}
For example, the \emph{batch-consistent sampling (BCS)} technique fixes the injected noise across all video frames~\citep{kwon2024visionxlhighdefinitionvideo, kwon2025solving}.
Another common approach, which we refer to as \emph{noise warping}, first extracts optical flow from measurements and uses it to warp the injected noise across all video frames~\citep{chang2024how, daras2024warped, deng2025infiniteresolution}.
As observed empirically, the dynamics of the injected noise translate into those of the generated video.
\rebuttal{More recently, video generative models have been used directly as priors. \citet{kwon2024seeing} use a pixel-space video diffusion model to see through dynamic scattering layers, LVTINO~\citep{spagnoletti2025lvtino} uses a latent video consistency model for zero-shot high-definition video restoration, and AVIS~\citep{kwon2026avis} speeds up video restoration with an autoregressive video diffusion model that restores the video chunk by chunk; LVTINO and AVIS build on video models pretrained on large natural-video corpora. These methods mostly target natural videos, whose dynamics are largely preserved in the measurements or for which training data are abundant, whereas we learn priors from several hundred scientific simulations (Sec.~\ref{sec:baselines} compares with a natural-video model adapted to our data).}

\paragraph{\rebuttal{Spatiotemporal priors beyond diffusion models}}
\rebuttal{Learned spatiotemporal priors predate video diffusion models: video denoisers such as DVDnet~\citep{tassano2019dvdnet}, ViDeNN~\citep{claus2019videnn}, and FastDVDnet~\citep{tassano2020fastdvdnet} exploit spatial and temporal correlations across neighboring frames, and have been used as plug-and-play priors for video deblurring, super-resolution, and interpolation of missing pixels~\citep{monod2022video}. Unlike such denoisers, a diffusion prior yields multiple posterior samples rather than a single estimate. Other approaches avoid the need for clean training videos, which is attractive when such data are scarce. \citet{wang2024fully} train a dynamic MRI reconstruction network from undersampled measurements alone by exploiting spatiotemporal equivariances. For dynamic imaging with very long baseline interferometry (VLBI), StarWarps~\citep{bouman2018reconstructing} imposes a hand-crafted Gaussian Markov model on the video, and \kine~\citep{foschi2026kine} reconstructs videos of variable radio sources by fitting a neural-field representation of the video directly to interferometric observations. \ours~instead learns the spatiotemporal prior from simulated videos, so its results depend on how representative the simulations are.}

\paragraph{Plug-and-play diffusion priors for inverse problems}
Plug-and-play diffusion priors (PnPDP) constitute a family of methods that leverage diffusion models as priors for solving inverse problems.
The goal is to draw samples from the posterior distribution \(p(\rvx_0|\rvy) \,\propto\, p(\rvy|\rvx_0) p(\rvx_0) \)
where the likelihood \(p(\rvy|\rvx_0)\) can be derived from knowledge of the physical system, and the prior $p(\rvx_0)$ is characterized by a diffusion model.
Various formulations have been considered in the existing literature, including measurement guidance~\citep{chung2023diffusion,rout2023solving,kawar2022denoising,song2023pseudoinverseguided}, variable splitting~\citep{coeurdoux2023plugandplay, li2024decoupleddataconsistencydiffusion, song2024solvinginverseproblemslatent, xu2024provably, wu2024principled,zhang2024improving}, variational Bayes~\citep{mardani2023variational, feng2023score, feng2024variational}, and Sequential Monte Carlo (SMC)~\citep{wu2023practical, trippe2023diffusion, cardoso2023monte, dou2024diffusion}.
\rebuttal{Because the prior is decoupled from the likelihood, a prior trained for one data domain can in principle be combined with different forward models without retraining.}

\paragraph{Diffusion models as image and video priors}
Latent diffusion models (LDMs)~\citep{rombach2022high} offer a way to reduce the computational cost otherwise necessary to model the original high-dimensional data distribution.
LDMs generate an efficient, low-dimensional latent representation $\rvz_0$ of data $\rvx_0$ with a pretrained perceptual compression encoder $\mathcal E$ and decoder $\mathcal D$, which satisfy $\rvz_0=\mathcal E(\rvx_0)$ and $\mathcal D(\rvz_0)\approx \rvx_0$.
\rebuttal{Diffusion models have been extended to videos, typically by factorizing the network into spatial and temporal layers~\citep{ho2022videodiffusionmodels}.
Many recent methods obtain a video model by fine-tuning a pretrained IDM on videos~\citep{wu2023tuneavideooneshottuningimage,blattmann2023alignlatentshighresolutionvideo,blattmann2023stablevideodiffusionscaling}, using techniques such as identity initialization of the added temporal layers~\citep{singer2023makeavideo} and learned blending of spatial and temporal features~\citep{blattmann2023alignlatentshighresolutionvideo}. Video diffusion models also benefit from joint training on images and videos, with the temporal layers bypassed for images~\citep{ho2022videodiffusionmodels, ho2022imagenvideo}.
\ours~adopts several of these techniques (Tab.~\ref{tab:prior-sources}) to build priors for scientific video inverse problems.}

\section{Method}
\label{sec:method}

In this section, we describe \ours, which solves video inverse problems with \textbf{S}patio\textbf{Te}mporal diffusion \textbf{P}riors.
\Figref{fig:method} illustrates the conceptual difference between our approach and existing methods for video inverse problems.
Designed mainly for restoration problems on natural videos, most existing methods rely on IDMs to incorporate a spatial prior and enforce temporal consistency by injecting correlated noise in the diffusion process.
As we will show in the experiments, these techniques do not work well on the ill-posed scientific problems we consider.
Our method instead adopts a whole-video formulation (\Secref{sec:basic-formulation}) and simultaneously handles spatial and temporal dimensions within a PnPDP framework (\Secref{sec:vdm-sampler}).
\rebuttal{The spatiotemporal diffusion prior (\Secref{sec:vdm-prior}) is built with standard video-generation techniques and learns temporal information directly from data, which removes} the need for designing task-specific temporal heuristics.

\subsection{General approach}
\label{sec:basic-formulation}

We consider the general formulation of video inverse problems where the goal is to recover a moving target \(\rvx_0\) from the time-varying measurements
\begin{equation}\label{eq:forward-model}
    \rvy = \mathcal{A}(\rvx_0) + \rvn.
\end{equation}
Here \(\mathcal{A}(\cdot)\) is the forward model, which could be nonlinear, and \(\rvn\) is the measurement noise.
In many scientific problems, such as black hole video reconstruction, substantial spatiotemporal information of \(\rvx_0\) is lost in the measurement process due to the ill-posed and potentially nonlinear nature of \(\mathcal{A}(\cdot)\), necessitating a prior on both the spatial and temporal dimensions of \(\rvx_0\) for meaningful recovery.

Instead of taking a per-frame processing approach using IDMs, we directly learn the video distribution \(p(\rvx_0)\) from data and draw samples from the posterior distribution \(p(\rvx_0|\rvy) \,\propto\, p(\rvy|\rvx_0) p(\rvx_0)\) via the PnPDP framework.
This enables us to deal with more general problems in which the extraction of temporal information directly from measurements is hard.

\paragraph{Solve video inverse problems in latent space.}
While the likelihood \(p(\rvy|\rvx_0)\) can be derived from \Eqref{eq:forward-model}, it is often challenging to characterize the prior distribution \(p(\rvx_0)\) for videos due to its high dimensionality and potentially limited number of samples for training.
To overcome these challenges, we follow latent diffusion models~\citep{rombach2022high} and impose a spatiotemporal prior in the latent space.
Assuming that the set of likely \(\rvx_0\)'s is in the range of a decoder $\mathcal{D}$, we have that \(\exists \, \rvz_0\) s.t. \(\rvx_0 = \mathcal{D}(\rvz_0)\) and can thus rewrite \Eqref{eq:forward-model} as:
\begin{align}
\label{eq:latent-forward-model}
    \rvy = \mathcal{A}(\mathcal{D}(\rvz_0)) + \rvn.
\end{align}
It follows that the posterior \(p(\rvx_0|\rvy)\) is the pushforward of the latent posterior \(p(\rvz_0|\rvy)\) through \(\mathcal{D}\), so it suffices to first generate latent samples from \(p(\rvz_0|\rvy)\) and then decode them by \(\mathcal{D}\).

\subsection{Decoupled Annealing Posterior Sampling}
\label{sec:vdm-sampler}

{\ours} leverages the PnPDP framework to sample the posterior distribution given by~\Eqref{eq:latent-forward-model}.
In this work, we employ the Decoupled Annealing Posterior Sampling (DAPS) method~\citep{zhang2024improving}, which has demonstrated strong performance on several scientific inverse problems, including black hole imaging~\citep{zheng2025inversebench}.
It is also easily compatible with latent diffusion models, making it a suitable choice for our purpose.

DAPS samples the target latent posterior $p(\rvz_0|\rvy)$ by sequentially sampling $p(\rvz_t|\rvy)$ from $t=T$ to $t=0$.
To do so, DAPS first generates a sample from \(p (\rvz_T|\rvy) \approx \mathcal{N}(\vzero, \sigma_{\max}^2 \mI)\) and sequentially draws a sample from $p(\rvz_{t_{i-1}}|\rvy)$ given a sample from \( p(\rvz_{t_{i}}|\rvy) \) for $i=N,...,1$ based on a time schedule $\{t_i\}_{i=1}^{N}$.
As shown by Proposition 1 of \citep{zhang2024improving}, this is possible if one can sample from
\begin{align*}
    p(\rvz_0|\rvz_t, \rvy) = \frac{p(\rvy|\rvz_0, \rvz_t)p(\rvz_0|\rvz_t)}{p(\rvy|\rvz_t)} \,\propto\, p(\rvy|\rvz_0)p(\rvz_0|\rvz_t).
\end{align*}
Since we have access to the gradient of this unnormalized distribution, we can sample it using MCMC methods, such as Langevin Monte Carlo (LMC)~\citep{welling2011bayesian} and Hamiltonian Monte Carlo (HMC)~\citep{betancourt2015hamiltonian}.
After obtaining \( \rvzhatzero \sim p(\rvz_0|\rvz_{t_i}, \rvy) \), we then sample from \( p(\rvz_{t_{i-1}}|\rvy) \) by sampling $\rvz_{t_{i-1}} \sim \mathcal{N}(\rvzhatzero, \sigma_{t_{i-1}}^2 \mI)$ due to Proposition 1 of \citep{zhang2024improving}.
\rebuttal{Apart from using HMC for the MCMC step, we use DAPS without modification; Appendix~\ref{app-pipeline} gives the pseudocode (\Algref{alg}) and further details.}

\subsection{Training spatiotemporal diffusion priors}
\label{sec:vdm-prior}

\rebuttal{We construct the prior in three steps (Tab.~\ref{tab:prior-sources}); Alg.~\ref{alg:train} summarizes the training procedure, and \Secref{sec:baselines} gives the hyperparameters and training cost.}

\begin{table}[t]
    \rebuttalon
    \centering
    \footnotesize
    \caption{\rebuttal{\textbf{Components of \ours~and their sources.} \ours~combines established techniques from video generation with an existing posterior sampler. In Tab.~\ref{tab:main-table}, BCS and \ours~(video only) ablate the temporal layers and the image-based training, respectively.}}
    \label{tab:prior-sources}
    \setlength{\tabcolsep}{3.5pt}
    \begin{tabular}{@{}lll@{}}
        \toprule
        & \textbf{Component} & \textbf{Source} \\
        \midrule
        Step 1 & Latent diffusion with a per-frame 2D VAE & LDM~\citep{rombach2022high} \\
        Step 2 & Factorized spatial and temporal layers & VDM~\citep{ho2022videodiffusionmodels} \\
               & Learned blending of spatial and temporal outputs & Video LDM~\citep{blattmann2023alignlatentshighresolutionvideo} \\
        Step 3 & Joint training on videos and independent images & VDM~\citep{ho2022videodiffusionmodels} \\
               & Temporal layers bypassed for image inputs & Imagen Video, SVD~\citep{ho2022imagenvideo, blattmann2023stablevideodiffusionscaling} \\
        Sampling & Decoupled annealing posterior sampling & DAPS~\citep{zhang2024improving} \\
        \bottomrule
    \end{tabular}
\end{table}

\paragraph{Step 1: Training a latent diffusion model (LDM) as an image prior}
\rebuttal{We start by training a VAE~\citep{kingma2022autoencodingvariationalbayes} on an image dataset~$D_\text{image}$ with the standard $L_1$ reconstruction loss and a scaled KL divergence~\citep{burgess2018understandingdisentanglingbetavae}:
\begin{equation}\label{eq:vae-loss}
    \mathcal{L}_{\text{VAE}} = \mathbb{E}_{\rvx_0\sim D_{\text{image}}} \Big[ \mathbb{E}_{\rvz_0\sim q_{\phi}(\rvz_0 \mid \rvx_0)} \tfrac{1}{d} \| \mathcal{D}(\rvz_0) - \rvx_0 \|_1 + \beta_\text{KL}\, D_{\text{KL}} \big( q_{\phi}(\rvz_0 \mid \rvx_0) \,\|\, \mathcal{N}(\vzero, \mI) \big) \Big],
\end{equation}
where $d$ is the number of pixels and $q_\phi(\rvz_0\mid\rvx_0)$ is a Gaussian with diagonal covariance whose mean and standard deviation are given by the encoder $\mathcal E(\rvx_0)$.
The KL weight $\beta_\text{KL}$ is set to much less than 1 to prevent excessive regularization of the latent space; its value balances reconstruction accuracy against the convergence speed of diffusion training.
Once $\mathcal E$ and $\mathcal D$ are trained, we fix their parameters and train a 2D UNet noise-prediction network $\boldsymbol{\epsilon}_{\rvtheta}$ with the standard DDPM loss~\citep{ho2020denoising}, a reweighted form of denoising score matching~\citep{vincent2011aconnection}:
\begin{equation}\label{eq:idm-loss}
    \mathcal{L}_{\text{DM}}(\rvtheta) = \mathbb{E}_{\rvz_0,\, k,\, \boldsymbol{\epsilon}} \Big[ \big\| \boldsymbol{\epsilon}_{\rvtheta}\big(\sqrt{\bar\alpha_k}\,\rvz_0 + \sqrt{1-\bar\alpha_k}\,\boldsymbol{\epsilon},\, k\big) - \boldsymbol{\epsilon} \big\|^2 \Big],
\end{equation}
where $\rvz_0\sim q_\phi(\rvz_0\mid\rvx_0)$ with $\rvx_0\sim D_\text{image}$, $k\sim\mathcal{U}\{1,\dots,K\}$, $\boldsymbol{\epsilon}\sim\mathcal{N}(\vzero,\mI)$, and $\bar\alpha_k$ follows the linear noise schedule of DDPM with $K=1000$ steps. For sampling, $\boldsymbol{\epsilon}_{\rvtheta}$ gives the score of the equivalent variance-exploding diffusion (Appendix~\ref{app-pipeline}).}
Despite recent progress in 3D spatiotemporal encoders and decoders \citep{yang2024cogvideoxtexttovideodiffusionmodels,wu2024improvedvideovaelatent,chen2024odvaeomnidimensionalvideocompressor}, we opt for a 2D spatial encoder and decoder to process each frame independently.
This choice is due to efficiency considerations, as the decoder \( \mathcal{D} \) will be called repeatedly in DAPS during inference. 

\begin{figure}[t]
    \centering
    \begin{minipage}[c]{0.44\linewidth}
        \centering
        \includegraphics[width=\linewidth]{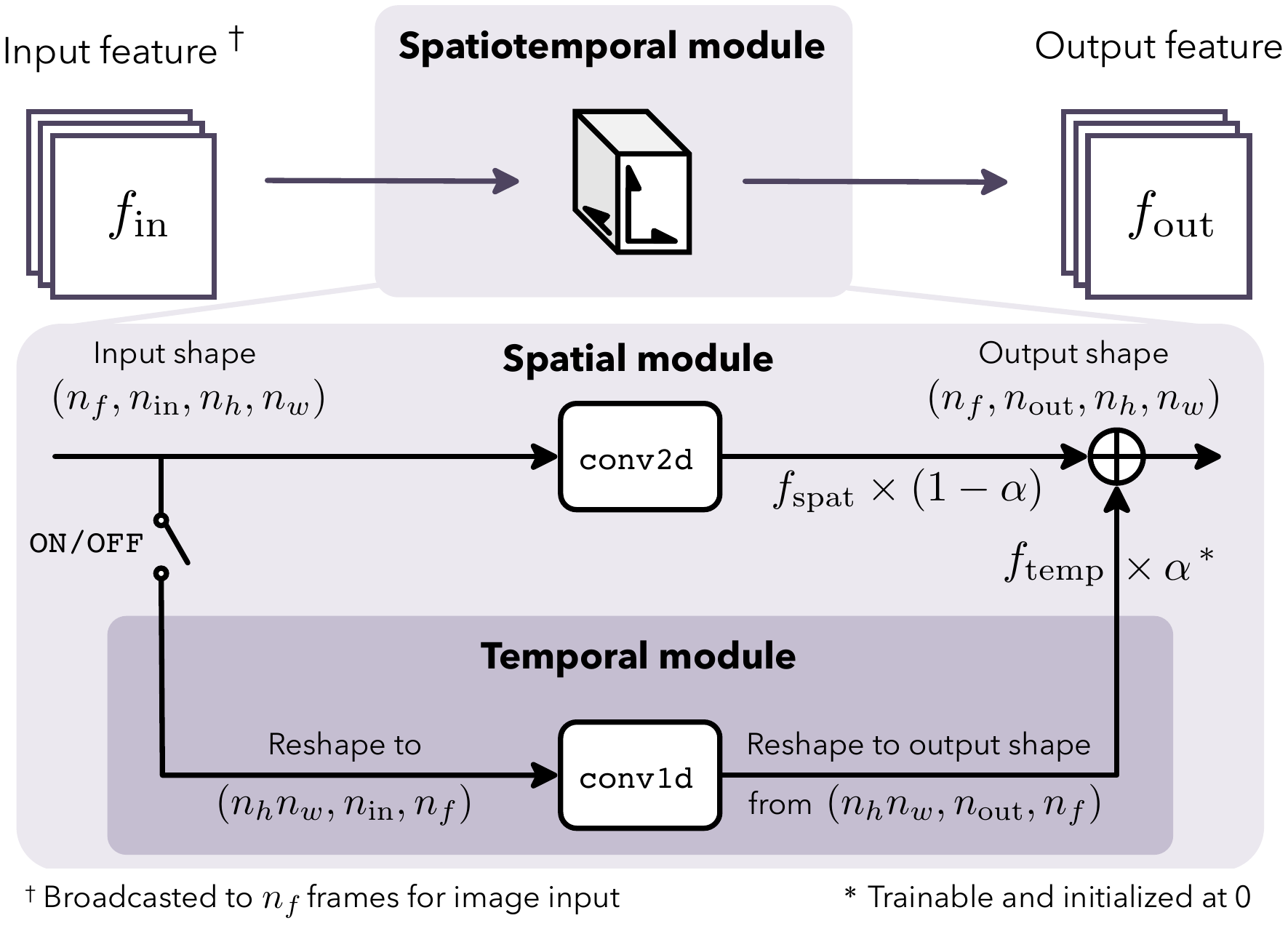}
    \end{minipage}\hfill
    \begin{minipage}[c]{0.53\linewidth}
    \caption{
    \textbf{Architecture of the spatiotemporal module.} 
    \rebuttal{Given an image diffusion UNet, we add a temporal module with an \texttt{ON/OFF} switch to each 2D spatial module and blend the outputs of the two branches with a learnable weight $\alpha$; for image inputs, the switch is \texttt{OFF} ($\alpha=0$). The figure is schematic; Appendix~\ref{sec:app-training-details} gives the exact parameterization and initialization of $\alpha$.} The number of frames, height, and width are denoted by \(n_f\), \(n_h\), and \(n_w\), respectively. The numbers of channels for input features (\(f_\text{in}\)) and output features (\(f_\text{out}\)) are denoted by \(n_\text{in}\) and \(n_\text{out}\), respectively.}
    \label{fig:st-module}
    \end{minipage}
\end{figure}

\paragraph{Step 2: Turning an image prior into a spatiotemporal prior}
Upon obtaining an LDM as an image prior, we use a spatiotemporal UNet architecture for $\boldsymbol{\epsilon}_{\rvtheta}$, which gives the time-dependent video score function \( \rvs_{\rvtheta}(\rvz_t; \sigma_t) \approx \nabla_{\rvz_{t}}\log p(\rvz_t;\sigma_t) \), leveraging recent advancements in video generation \citep{wang2023modelscopetexttovideotechnicalreport,ho2022videodiffusionmodels}.
The key component in the architecture is the spatiotemporal module for 3D modeling, as illustrated in \Figref{fig:st-module}.
Given the IDM with a 2D UNet from Step 1, we add a temporal module to each 2D spatial module within the UNet.
Specifically, for an input feature \( f_\text{in} \), let \( f_\text{out} \) be the output of the spatiotemporal module, with \( f_\text{spat} \) and \( f_\text{temp} \) representing the outputs of the spatial and temporal branches, respectively.
These features are combined using an $\alpha$-blending mechanism:  
\begin{equation}
    f_\text{out} = (1-\alpha) \cdot f_\text{spat} +  \alpha \cdot f_\text{temp}.
\end{equation}
\rebuttal{Here $\alpha \in (0,1)$ is a learnable blending weight in each spatiotemporal module; Appendix~\ref{sec:app-training-details} gives its parameterization and initialization.}
This design allows us to inherit the weights of the 2D spatial modules from the IDM, reducing training time.
Additionally, by factorizing the 3D module into a 2D spatial module and a 1D temporal module, the spatiotemporal UNet has marginal computational overhead
compared to the original 2D UNet, striking a good balance between model capacity and efficiency.

\paragraph{Step 3: Image-video joint fine-tuning}
To train on both images and videos, we use an \texttt{ON/OFF} switch signal in the spatiotemporal module.
When the switch is set to \texttt{OFF} (indicating image inputs), the temporal module is disabled (or equivalently \( \alpha = 0 \)).
This ensures that \( f_\text{out} = f_\text{spat} \) and reduces the spatiotemporal module to the original 2D spatial module, which processes each frame independently.  
\rebuttal{We initialize the weights of the spatial modules from the IDM of Step 1 and fine-tune all parameters of the spatiotemporal UNet. Each iteration takes a gradient step on a batch of videos and, with probability $p_\text{img}$, an additional step on a batch of pseudo-videos of independent images with the switch set to \texttt{OFF} (\Algref{alg:train}); both steps use the loss in \Eqref{eq:idm-loss} on the per-frame latents. The image steps help the spatiotemporal UNet retain the spatial capabilities of the IDM, in line with prior findings that joint image-video training improves sample quality~\citep{ho2022videodiffusionmodels} and mitigates forgetting of image-domain knowledge~\citep{wang2023modelscopetexttovideotechnicalreport}.}

\begin{algorithm}[t]
    \rebuttalon
    \caption{\rebuttal{Training the spatiotemporal diffusion prior of \ours}}
    \label{alg:train}
    \begin{algorithmic}[1]
        \Require image dataset $D_\text{image}$, video dataset $D_\text{video}$, KL weight $\beta_\text{KL}$, image-step probability $p_\text{img}$
        \State Train $\mathcal E, \mathcal D$ on $D_\text{image}$ by minimizing $\mathcal L_\text{VAE}$ (\Eqref{eq:vae-loss}); freeze $\mathcal E, \mathcal D$ \Comment{Step 1}
        \State Train a 2D UNet $\boldsymbol{\epsilon}_\rvtheta$ on the latents of $D_\text{image}$ by minimizing $\mathcal L_\text{DM}$ (\Eqref{eq:idm-loss})
        \State Add a temporal module with a learnable blending weight $\alpha$ to each 2D spatial module of $\boldsymbol{\epsilon}_\rvtheta$ \Comment{Step 2}
        \Repeat \Comment{Step 3}
            \State $\rvx_0 \sim D_\text{video}$, switch \texttt{ON}; gradient step on $\mathcal L_\text{DM}$ (\Eqref{eq:idm-loss}) for all parameters
            \State W.p.\ $p_\text{img}$: $\rvx_0 \gets$ pseudo-video of images from $D_\text{image}$, switch \texttt{OFF}; gradient step on $\mathcal L_\text{DM}$
        \Until{the number of epochs in Tab.~\ref{tab:app-vdm-training} is reached}
        \State \Return spatiotemporal prior $\boldsymbol{\epsilon}_\rvtheta$ and decoder $\mathcal D$
    \end{algorithmic}
\end{algorithm}

\section{Experiments}
\label{sec:experiment}

\subsection{Tasks and setup}
\label{sec:exp_setup}

\rebuttal{We consider black hole video reconstruction~\citep{akiyama2022firstIII} as our primary task and dynamic MRI~\citep{gamper2008compressed} as a secondary application.}
Although both are scientific imaging tasks, they have significantly different characteristics. Black hole video reconstruction involves simple spatial structures (usually a ring structure) but complex temporal dynamics that obey physical constraints.
In contrast, dynamic MRI requires higher spatial fidelity with relatively simpler temporal dynamics, such as periodic heartbeat motion.

\paragraph{Black hole video reconstruction}
We consider the problem of observing the Sagittarius A$^\ast$ black hole using the Event Horizon Telescope (EHT) array in 2017~\citep{akiyama2022firstIII}.
The measurements form a 1,856-dimensional vector for an underlying black hole video of 64 frames, each with a spatial resolution of 256$\times$256 pixels (about 4.2 million unknowns), capturing the black hole's dynamics during the 100-minute observation period.
These measurements are given by calculating the closure quantities based on the complex visibilities (a detailed description of the problem is available in Appendix~\ref{sec:bh-exp-details}).
For the training dataset, we employ the \emph{general relativistic magnetohydrodynamic (GRMHD)} simulations~\citep{wong2022patoka} of Sagittarius A$^\ast$, covering various black hole models and observational conditions. \rebuttal{The dataset comprises 648 simulated black hole videos and, separately, 50,000 simulated black hole images drawn i.i.d.\ from the same distribution (Appendix~\ref{sec:app-training-details}).}

\begin{table}[t!]
\scriptsize
\centering
\caption{\textbf{Quantitative results on black hole video reconstruction and dynamic MRI.}
\rebuttal{Mean and standard deviation (in parentheses) over 10 test videos; for black holes, we use for each method and video the sample with the highest PSNR out of 20 (Sec.~\ref{sec:metrics}). The best result is in bold and the second best is underlined. On black holes, the noise-warping baselines are given oracle optical flow computed from the ground-truth videos, so their numbers represent an upper bound on their achievable performance. Tab.~\ref{tab:app-paired} reports paired differences, and Appendix~\ref{sec:app-fvd} reports FVD.}}
\label{tab:main-table}
\resizebox{\linewidth}{!}{
\begin{tabular}{lccc!{\vrule}cc!{\vrule}c}
    \toprule
    \multicolumn{7}{c}{\textbf{Task: Black hole}}\\
    \midrule
     \textbf{Methods} & \textbf{PSNR ($\uparrow$)} & \textbf{SSIM ($\uparrow$)} & \textbf{LPIPS ($\downarrow$)} & \textbf{d-PSNR ($\uparrow$)} & \textbf{d-SSIM ($\uparrow$)} & \textbf{Data Misfit ($\downarrow$)} \\ \midrule

BIS~\citep{kwon2025solving}  & 23.79 (1.41) & 0.718 (0.047) & 0.179 (0.031) & 29.26 (1.51) & 0.938 (0.015) & 1.719 (1.277)\\
    BCS~\citep{kwon2025solving} & 27.66 (2.04) & 0.816 (0.053) & 0.124 (0.040) & \underline{41.71} (1.99) & \textbf{0.979} (0.008) & 1.426 (0.784)\\
    Bilinear~\citep{chang2024how} & 26.11 (2.05) & 0.718 (0.067) & 0.151 (0.044) & 33.14 (2.33) & 0.958 (0.013) & 1.384 (0.742) \\
    Bicubic~\citep{chang2024how} & 25.68 (1.88) & 0.730 (0.058) & 0.163 (0.041) & 31.70 (1.88) & 0.952 (0.015) & 1.736 (1.259)  \\
    Nearest~\citep{chang2024how} & 25.29 (1.80) & 0.754 (0.059) & 0.164 (0.042) & 30.76 (1.75) & 0.943 (0.017) & 1.691 (1.020)  \\
    $\int$-noise~\citep{deng2025infiniteresolution} & 24.90 (1.52) & 0.745 (0.058) & 0.163 (0.034) & 31.87 (1.93) & 0.945 (0.017) & 1.655 (1.020) \\
    GP-Warp~\citep{daras2024warped} & 23.98 (1.28) & 0.721 (0.043) & 0.176 (0.029) & 29.21 (1.24) & 0.938 (0.014) & 1.721 (1.225) \\
     \midrule
    \rebuttal{Wan2.1 + LoRA~\citep{wan2025wan,hu2022lora}} & \rebuttal{27.41 (2.61)} & \rebuttal{\underline{0.846} (0.094)} & \rebuttal{\underline{0.112} (0.051)} & \rebuttal{40.87 (2.13)} & \rebuttal{0.975 (0.009)} & \rebuttal{\underline{1.116} (0.134)} \\
     \midrule
    \ours~(video only) & \underline{28.71} (1.81) & 0.802 (0.079) & 0.120 (0.041) & 41.41 (2.39) & 0.975 (0.011) & 1.124 (0.136) \\
    \ours~(image-video joint) & \textbf{30.28} (2.71) & \textbf{0.865} (0.063) & \textbf{0.095} (0.039) & \textbf{42.09} (2.63) & \underline{0.976} (0.011) & \textbf{1.114} (0.154) \\
    \midrule
    \multicolumn{7}{c}{\textbf{Task: MRI ($8\times$)}}\\
    \midrule
     \textbf{Methods} & \textbf{PSNR ($\uparrow$)} & \textbf{SSIM ($\uparrow$)} & \textbf{LPIPS ($\downarrow$)} & \textbf{d-PSNR ($\uparrow$)} & \textbf{d-SSIM ($\uparrow$)} & \textbf{Data Misfit ($\downarrow$)} \\ \midrule
    BIS~\citep{kwon2025solving}  & 35.04 (0.76) & 0.889 (0.016) & 0.100 (0.012) & 38.91 (1.13) & 0.918 (0.012) & 9.203 (0.698) \\
  BCS~\citep{kwon2025solving} & 35.43 (0.97) & 0.893 (0.016) & 0.099 (0.012) & 39.91 (1.09) & 0.931 (0.011) & 9.225 (0.693) \\
    Bilinear~\citep{chang2024how}  & 35.30 (0.96) & 0.896 (0.016) & 0.098 (0.012) & 39.87 (1.14) & 0.931 (0.011) & 9.157 (0.670) \\
    Bicubic~\citep{chang2024how}  & 35.31 (0.91) & 0.896 (0.016) & 0.098 (0.011) & 40.11 (1.10) & 0.934 (0.010) & 9.189 (0.653) \\
    Nearest~\citep{chang2024how}  & 34.87 (0.90) & 0.895 (0.018) & 0.099 (0.013) & 40.09 (1.14) & 0.933 (0.011) & 9.188 (0.656) \\
    $\int$-noise~\citep{deng2025infiniteresolution} & 35.55 (1.03) & 0.892 (0.020) & 0.099 (0.014) & 39.77 (1.36) & 0.929 (0.015) & 9.208 (0.693) \\
    GP-Warp~\citep{daras2024warped} & 34.49 (0.65) & 0.886 (0.016) & 0.102 (0.012) & 38.71 (1.14) & 0.916 (0.013) & 9.209 (0.659) \\
   \midrule
    \ours~(video only)  & \underline{37.00} (1.46) & \underline{0.927} (0.019) & \underline{0.086} (0.013) & \underline{43.50} (2.70) & \underline{0.963} (0.015) & \underline{8.817} (0.650) \\
    \ours~(image-video joint) & \textbf{39.38} (1.16) & \textbf{0.951} (0.009) & \textbf{0.078} (0.011) & \textbf{44.86} (1.92) & \textbf{0.974} (0.006) & \textbf{8.753} (0.603)\\
    \bottomrule
\end{tabular}}
\end{table}

\paragraph{Dynamic MRI}
We consider a standard compressed sensing MRI setup with two acceleration scenarios: 8$\times$ acceleration with 12 auto-calibration signal (ACS) lines and 6$\times$ acceleration with 24 ACS lines.
Further details are provided in Appendix~\ref{sec:mri-exp-details}.
We utilize the publicly available cardiac cine MRI dataset from the CMRxRecon Challenge 2023~\citep{wang2024cmrxrecon} for training.
This dataset contains 3,324 fully sampled, ECG-triggered cardiac cine sequences from 300 patients, covering various canonical cardiac imaging views.
Each sequence is processed into a video consisting of 12 frames with a spatial resolution of 192$\times$192 pixels.
\rebuttal{Since no separate image set from the same distribution is available, we construct the image dataset by extracting the individual frames from the training videos, resulting in 39,888 images in total.
The image and video datasets therefore contain the same frames, so, unlike for black holes, the image model serves as an initialization rather than as pretraining on additional data.}

\paragraph{Evaluation protocol}
\rebuttal{For both tasks, all quantitative results are reported on 10 test videos that were drawn i.i.d.\ and held out at the outset; they appear in no training data, neither as videos nor as frames of the image datasets.} A separate held-out validation set of 3 videos, also disjoint from both the training and evaluation sets, is used solely for tuning the sampler hyperparameters (Tab.~\ref{tab:app-step-hyper}).

\subsection{Baselines and our method}
\label{sec:baselines}

Recall from Sec.~\ref{sec:method} that our method leverages a learned spatiotemporal prior to avoid explicitly estimating temporal dynamics from measurements at inference time.
To evaluate the effectiveness of this strategy, we compare our method to existing IDM-based approaches.
Based on how temporal information is incorporated, we categorize these baseline methods into two groups (Groups 1 and 2). \rebuttal{We also compare with a large pretrained video diffusion model adapted to our data (Group 3).}

\paragraph{Group 1: simple heuristics}
We consider two baselines: BIS~\citep{kwon2025solving} and BCS~\citep{kwon2025solving}. \emph{Batch independent sampling} (BIS) reconstructs each frame independently, while \emph{batch consistent sampling} (BCS) implicitly incorporates a static temporal prior to enforce consistency across frames.

\begin{figure}[t!]
    \centering
    \includegraphics[width=\linewidth]{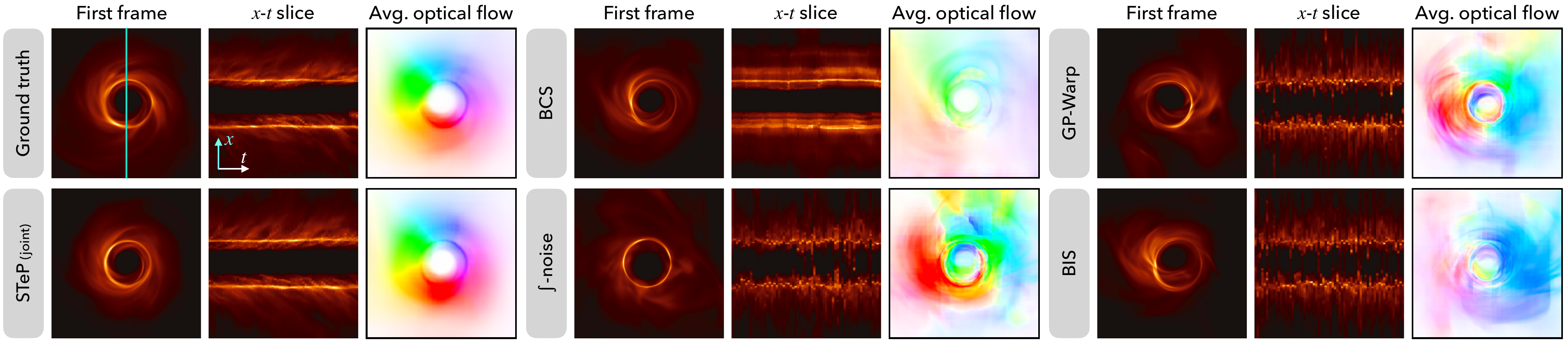}
    \caption{\textbf{Visual examples of \ours~(bottom left) and baselines for black hole video reconstruction.} To facilitate analysis of the reconstructed spatiotemporal structures, we present results in three ways: (1) the first frame to illustrate spatial fidelity, (2) an $x$-$t$ slice depicting temporal evolution of a vertical line to evaluate temporal consistency, and (3) the averaged optical flow visualized using the standard color wheel as implemented by~\citet{teed2020raftrecurrentallpairsfield} to assess spatiotemporal coherence jointly. The noise-warping baselines ($\int$-noise and GP-Warp) use oracle optical flow from the ground-truth video. \ours~matches the ground truth more closely than all baselines in all three views.}
    \label{fig:blackhole-imaging}
\end{figure}

\paragraph{Group 2: noise warping}
We include another line of baselines: $\int$-noise~\citep{chang2024how, deng2025infiniteresolution}, and GP-Warp~\citep{daras2024warped}. These baselines enforce temporal consistency by warping the noise using optical flow estimated directly from the measurements. We also include the simpler \emph{Bilinear}, \emph{Bicubic}, and \emph{Nearest} warping strategies from~\citet{chang2024how}. For the black hole video reconstruction task, a direct inversion from the sparse measurement vector is infeasible. Thus, to demonstrate the upper-bound performance of these methods, we leverage the ground truth video as an oracle to derive the optical flow. For the dynamic MRI task, we utilize a naive inversion obtained via inverse Fourier transformation to estimate the optical flow. We follow the methodology of~\citet{chang2024how}, leveraging a pretrained model in~\citet{teed2020raftrecurrentallpairsfield} to extract optical flow.

\paragraph{\rebuttal{Group 3: an adapted video model}}
\rebuttal{We adapt Wan2.1-T2V-1.3B~\citep{wan2025wan}, a general text-to-video diffusion model pretrained on large-scale natural videos (distilled Wan models are the priors of LVTINO and AVIS): we fine-tune the base model with rank-64 LoRA~\citep{hu2022lora} on the same 648 videos as \ours~(video only) and use it in the same DAPS sampler (Appendix~\ref{sec:baseline-implementation}).}

\paragraph{\rebuttal{Domain-specific methods}}
\rebuttal{Our simulated observations are designed to test learned priors under extremely sparse measurements: they span about 100 minutes, less than the EHT observation of Sagittarius A$^\ast$ on 2017 April 11, which is already near the limit of the data needed for imaging. Dynamic VLBI imaging methods that do not rely on a learned prior, such as StarWarps~\citep{bouman2018reconstructing} and \kine~\citep{foschi2026kine}, target observations with longer time coverage. We therefore do not include them as quantitative baselines, since a meaningful comparison would require adapting them to this regime.}

\paragraph{Ours: two variants of \ours} \rebuttal{We evaluate two variants differentiated by their spatiotemporal priors: \ours~(video only), whose spatiotemporal UNet is trained from scratch on videos only (with the same VAE), and \ours~(image-video joint), or \ours~(joint) for short, which initializes from the IDM of Step 1 (\Secref{sec:vdm-prior}) and undergoes joint image-video fine-tuning.}
To make a fair comparison, we run all the diffusion-based baselines with DAPS~\citep{zhang2024improving}. The detailed implementation of each baseline can be found in Appendices~\ref{app-pipeline} and~\ref{sec:baseline-implementation}.
Importantly, all Group 1 and Group 2 baselines use the \emph{same} image diffusion model checkpoint from which \ours~(image-video joint) is fine-tuned, and are run with the same DAPS solver; they differ from \ours~only in whether the prior models temporal structure and in how noise is coupled across frames. BCS can therefore be read as an ablation of \ours~in which the spatiotemporal modules are removed and temporal consistency is instead imposed by sharing noise across frames, so that the gap between BCS and \ours~(image-video joint) isolates the contribution of the learned spatiotemporal prior.

\paragraph{\rebuttal{Training details}}
\rebuttal{For both tasks, we use $\beta_\text{KL}=10^{-6}$ and, for \ours~(joint), $p_\text{img}=0.2$ (\Secref{sec:vdm-prior}). All stages use Adam with a learning rate of $2\times10^{-4}$ and keep an exponential moving average (EMA) of the weights, which is used in the next stage and for sampling. For black holes (dynamic MRI), we train the VAE, the IDM, and the spatiotemporal UNet for 25 (10), 200 (50), and 500 (300) epochs, which takes 4.5 (8.9), 5.5 (3.8), and 13.7 (22.8) hours on one A100 GPU. Tab.~\ref{tab:app-vdm-training} and Appendix~\ref{sec:app-training-details} give the remaining details.}

\subsection{Metrics}
\label{sec:metrics}

We evaluate our results on three aspects: (1) spatial similarity, (2) temporal consistency, and (3) measurement data fit.
\rebuttal{For black hole video reconstruction, we draw 20 samples per test video for every method and report the metrics of the sample with the highest PSNR with respect to the ground truth; for dynamic MRI, we report a single sample per test video.}

\paragraph{Spatial} Spatial similarity is assessed by calculating Peak Signal-to-Noise Ratio (PSNR), Structural Similarity Index Measure (SSIM)~\citep{wang2004image}, and Learned Perceptual Image Patch Similarity (LPIPS)~\citep{zhang2018lpips}. Metrics are computed per frame and averaged across all frames. We utilize implementations from \texttt{piq}~\citep{kastryulin2022piq}, normalizing images to the range $[0,1]$. For grayscale images, frames are replicated across three channels before computing LPIPS scores.

\paragraph{Temporal} Temporal consistency is quantified using delta-based PSNR (d-PSNR) and delta-based SSIM (d-SSIM), measuring the similarity of normalized differences between consecutive frames, with results averaged over all frames.
Note that delta-based metrics can favor temporally static reconstructions; we therefore interpret them together with the $x$-$t$ analyses in Sec.~\ref{sec:main-results}.
\rebuttal{We report the Fréchet Video Distance (FVD)~\citep{unterthiner2018towards} only in Appendix~\ref{sec:app-fvd}, because with 10 videos the estimate is strongly biased and its features, learned from natural videos, are not designed for black hole or cardiac data.}

\paragraph{Data fit} Lastly, we report measurement data fit using task-specific metrics. For the black hole video reconstruction task, we employ the unified average $\bm{\tilde\chi}^2$ statistic (defined in \Eqref{app:eq-chisquare-metrics} in Appendix), where values closer to 1 indicate better data fidelity. For dynamic MRI, we measure data misfit by computing the mean squared error in measurement space.

\paragraph{\rebuttal{Evaluating posterior samples}}
\rebuttal{\ours~draws approximate posterior samples, and we do not assess their calibration in this work (Sec.~\ref{app:limitation}). As is common for diffusion-based inverse problem solvers~\citep{chung2023diffusion, zhang2024improving, zheng2025inversebench}, we use reconstruction metrics against the ground truth and measurement data fit as surrogates for posterior quality, and we examine sample diversity qualitatively (\Secref{sec:main-results}).}

\subsection{Main results}
\label{sec:main-results}

\paragraph{\ours~reconstructs videos with better spatial and temporal coherence.}
Tab.~\ref{tab:main-table} reports the results for both tasks, and Fig.~\ref{fig:blackhole-imaging} and Fig.~\ref{fig:dynamic-mri} (Appendix~\ref{sec:app-more-res-visual}) show reconstructions for black holes and dynamic MRI, respectively.
\ours~outperforms the IDM-based baselines in overall video quality across most metrics and generates video reconstructions with better spatiotemporal coherence.
Fig.~\ref{fig:blackhole-imaging} visualizes an $x$-$t$ slice, representing the temporal evolution (horizontal axis) of a spatial slice (marked by the cyan vertical line).
We find that the noise warping baselines fail to constrain temporal consistency, as indicated by the fluctuating ring diameters in the $x$-$t$ slices, while BCS provides almost static reconstructions.
This near-static behavior also helps explain why BCS attains a slightly higher d-SSIM than \ours~on black hole video reconstruction, which illustrates why delta-based metrics alone are not sufficient to assess temporal fidelity.
In contrast, \ours~exhibits closer alignment with the ground truth and improved temporal consistency.
This is further illustrated by the averaged optical flow visualization, highlighting that \ours~faithfully captures the underlying temporal dynamics. Similar trends can be observed in Fig.~\ref{fig:dynamic-mri} (Appendix~\ref{sec:app-more-res-visual}).

\paragraph{Noise warping is less effective on these scientific problems.}
We observe that the performance of noise-warping strategies in challenging scientific video inverse problems does not correlate with noise-warping accuracy.
For example, although the $\int$-noise approach has demonstrated superior noise-warping capability given optical flow (as shown in~\citet{chang2024how}), it has a lower PSNR than simpler strategies such as \emph{Bilinear} and \emph{Bicubic} interpolation on black hole video reconstruction and is not consistently better on the other metrics (Tab.~\ref{tab:main-table}). We attribute this discrepancy to (1) inaccuracies in the optical flow, and (2) the inherent difficulty of effectively manipulating noise in latent space through pixel space optical-flow-guided warping. 
Consequently, methods relying on precise optical flow and carefully designed warping strategies struggle with tasks requiring high temporal consistency, such as black hole video reconstruction.
In contrast, \ours~learns temporal structure from training videos, which avoids these limitations on the tasks considered here.

\paragraph{\rebuttal{\ours~produces diverse solutions that fit the data comparably well.}}
Because the measurements in black hole video reconstruction are extremely sparse, \ours~can produce multiple semantically diverse reconstructions, each visually plausible and consistent with measurement data. 
Since the true posterior distribution is unknown, directly quantifying mode coverage is infeasible.
Instead, we draw 100 \textit{i.i.d.} samples and cluster them based on spatial appearance and temporal dynamics.
As illustrated in Fig.~\ref{fig:teaser}, we identify three distinct modes: one aligns closely with the ground truth, while the others differ in rotation direction or spatial structure.
\rebuttal{All three modes fit the data comparably well (Tab.~\ref{app:teaser-data-misfit}). These samples illustrate that the data admit distinct explanations; we do not interpret the mode frequencies as calibrated probabilities (Sec.~\ref{app:limitation}).}

\subsection{\rebuttal{Analysis of the spatiotemporal prior}}
\label{sec:effectiveness}

\paragraph{Joint fine-tuning benefits both spatial and temporal consistency.}
As shown in Tab.~\ref{tab:main-table}, \ours~with an image-video jointly fine-tuned spatiotemporal prior outperforms both IDM-based baselines and \ours~(video only) on all metrics except d-SSIM on black holes, where the near-static BCS is slightly higher. \rebuttal{On black holes, it also outperforms Wan2.1 + LoRA on every metric. Compared with \ours~(video only), which is trained on the same videos, Wan2.1 + LoRA is better on SSIM, LPIPS, and data fit but worse on PSNR and d-PSNR, while its denoiser is about 10$\times$ larger and sampling is about 4$\times$ slower (Appendix~\ref{sec:baseline-implementation}).} To further illustrate these improvements, we provide a detailed visual comparison in Fig.~\ref{fig:detailed-comparison} (Appendix~\ref{sec:app-more-res-visual}).
In Fig.~\ref{fig:detailed-comparison} (a), for black holes, we analyze the temporal dynamics using averaged delta frames, which visualize the averaged differences between consecutive frames over a window.
We observe that delta frames from BCS remain largely unchanged as the temporal window expands, indicating limited temporal variation.
In contrast, \ours~(joint) effectively captures more complex and coherent temporal dynamics.
In Fig.~\ref{fig:detailed-comparison} (b), for dynamic MRI, we compare \ours~(joint) and \ours~(video only) with a focus on spatial fidelity.
Although their reconstructions have similar average optical flow, \ours~(joint) has fewer artifacts and hallucinations as pointed out by the arrows.
\rebuttal{Because the MRI image dataset consists of the video frames themselves, this improvement comes from the training procedure, i.e., initializing from the IDM and continuing to train on individual frames, rather than from additional data.}

\paragraph{\rebuttal{Training data size.}}
\rebuttal{As a data-scaling ablation, we train both variants of \ours~on random subsets of 25\%, 50\%, 75\%, and 100\% of the 648 black hole training videos, with the 50{,}000 training images, the total number of training iterations, and all other settings fixed (Tab.~\ref{tab:data-efficiency}). Reconstruction quality generally improves with the number of videos for both variants. Joint image-video training is better at every size, and the PSNR gap is larger with fewer videos (3.40~dB with 162 videos vs.\ 1.57~dB with all 648).}

\begin{table}[t]
    \rebuttalon
    \scriptsize
    \centering
    \caption{\rebuttal{\textbf{Data-scaling ablation on black hole video reconstruction.} Mean over the 10 test videos (best of 20 samples, as in Tab.~\ref{tab:main-table}) for priors trained on a random subset of the 648 training videos (number of videos in parentheses), with the same 50{,}000 images and the same total number of training iterations as the full-data prior (hence more epochs for smaller subsets). The 100\% row corresponds to Tab.~\ref{tab:main-table}.}}
    \label{tab:data-efficiency}
    \setlength{\tabcolsep}{3.5pt}
    \begin{tabular}{lcccccc!{\vrule}cccccc}
        \toprule
        & \multicolumn{6}{c!{\vrule}}{\textbf{\ours~(video only)}} & \multicolumn{6}{c}{\textbf{\ours~(image-video joint)}} \\
        \textbf{Videos} & PSNR$\uparrow$ & SSIM$\uparrow$ & LPIPS$\downarrow$ & d-PSNR$\uparrow$ & d-SSIM$\uparrow$ & Misfit$\downarrow$ & PSNR$\uparrow$ & SSIM$\uparrow$ & LPIPS$\downarrow$ & d-PSNR$\uparrow$ & d-SSIM$\uparrow$ & Misfit$\downarrow$ \\
        \midrule
        25\% (162) & 24.56 & 0.702 & 0.172 & 37.22 & 0.954 & 1.320 & 27.96 & 0.824 & 0.117 & 40.16 & 0.959 & 1.220 \\
        50\% (324) & 26.82 & 0.756 & 0.133 & 39.40 & 0.967 & 1.257 & 28.84 & 0.830 & 0.111 & 40.43 & 0.968 & 1.121 \\
        75\% (486) & 27.41 & 0.763 & 0.136 & 39.08 & 0.964 & 1.261 & 29.29 & 0.849 & 0.109 & 41.19 & 0.972 & 1.149 \\
        100\% (648) & 28.71 & 0.802 & 0.120 & 41.41 & 0.975 & 1.124 & 30.28 & 0.865 & 0.095 & 42.09 & 0.976 & 1.114 \\
        \bottomrule
    \end{tabular}
\end{table}

\paragraph{\rebuttal{Fine-tuning progress.}}
\rebuttal{Evaluating checkpoints of joint fine-tuning runs continued to 750 epochs (Fig.~\ref{fig:ablation-posterior}, Appendix~\ref{sec:app-more-res-visual}), we find that PSNR, d-PSNR, and data misfit all improve over the first few hundred epochs and then plateau.}

\section{Limitations}
\label{app:limitation}
The sampling cost of \ours~is relatively high (645~s and 48~GB per black hole video; Tab.~\ref{tab:app-step-hyper}) due to the requirement of backpropagation through decoder $\mathcal D$ in MCMC updates in \Algref{alg}. This creates a trade-off between the capability of the decoder and its computational cost. We leave MCMC updates in pixel space, or other ways to bypass backpropagation through the decoder, to future work.

\rebuttal{A second limitation concerns the interpretation of our posterior samples. Although \ours~recovers multiple distinct modes that fit the measurements comparably well (Fig.~\ref{fig:teaser}), we do not claim that these modes and their frequencies form a calibrated approximation of the posterior, which is intractable here; assessing calibration would require either a reference posterior on a simplified problem or coverage tests over many test observations. We therefore use reconstruction metrics as surrogates (\Secref{sec:metrics}), and the samples should not be used for quantitative uncertainty estimates without further validation.}

\rebuttal{A third limitation is that our priors are trained on simulations, and their behavior on out-of-distribution data, such as real observations that deviate from the simulations, remains to be evaluated. Each prior is also trained for one data domain; although it can in principle be paired with other forward models without retraining, we test this only for two MRI acceleration settings, which share the same type of forward operator.}

Finally, as with any method that relies on a strong learned prior, reconstructions may contain structure driven by the prior rather than by the measurements. This is a particular concern in scientific settings, where reconstructions inform downstream analysis. As a qualitative check, drawing multiple samples can reveal which features vary across them, and we recommend inspecting a set of samples rather than a single reconstruction.

\section{Conclusion}
\label{sec:conclusion}

\rebuttal{We presented {\ours}, which brings learned spatiotemporal diffusion priors to scientific video inverse problems. By extending a latent image diffusion model with temporal layers, fine-tuning it jointly on images and videos, and pairing it with an existing posterior sampler, {\ours} learns temporal structure from limited training videos instead of imposing it through heuristics. On black hole video reconstruction from sparse EHT measurements, it recovers spatiotemporal structure that image diffusion baselines miss, and the same recipe also improves dynamic MRI reconstructions. We hope these results encourage further work on learned video priors for scientific imaging and on validating the calibration of their posterior samples.}

\section*{Broader Impact Statement}
\label{app:broader-impacts}
\rebuttal{Our work concerns scientific video inverse problems in astronomy and medical imaging, where reconstructions can inform scientific or clinical decisions. The learned prior may introduce biases, and the posterior samples drawn by \ours~are approximate. Before such samples inform decisions or conclusions, their calibration and their behavior on out-of-distribution data (e.g., real observations that deviate from the simulations) must be validated; practitioners should therefore evaluate the reconstructions carefully and mitigate potential negative impacts.}

\section*{Acknowledgments}
This research is funded by NSF award 2048237 and NSF award 2034306. B.Z is supported by the Kortschak Scholars Fellowship. Z.W. is supported by the Amazon AI4Science fellowship. B.F. is supported by the Pritzker Award.

We thank Ben Prather, Abhishek Joshi, Vedant Dhruv, C.K. Chan, and Charles Gammie for the synthetic black hole images GRMHD dataset used here, generated under NSF grant AST 20-34306.

\bibliography{main}
\bibliographystyle{tmlr}

\clearpage
\appendix

\section{Detailed implementation of \ours}
\label{app-pipeline}
Here, we summarize \suppchange{the sampling procedure of \ours} for solving video inverse problems in \Algref{alg}.
\begin{algorithm*}[ht]
\caption{\textbf{\ours}: \suppchange{posterior sampling for} video inverse problems with \suppchange{a} \textbf{S}patio\textbf{Te}mporal \textbf{P}rior}
\label{alg}
\begin{algorithmic}[1]
\Require Discretization time steps $\{t_i\}_{i=1}^{N}$ where $t_0=0$ and $t_N=T$, noise schedule $\sigma_t$, likelihood $p(\rvy \mid \cdot)$ with measurements $\rvy$, HMC step size $\eta$ and damping factor $\gamma$, number of HMC updates $M$, pretrained latent score function $\rvs_\rvtheta(\rvz; \sigma) \approx \nabla_\rvz \log p\left(\rvz; \sigma\right)$ with image decoder $\mathcal{D}$.
\State $\rvz_{t_N} \sim \mathcal{N}(\vzero, \sigma_{t_N}^2 \mI)$\Comment{Initialization}
\For{$i = N, ..., 1$}
    \State $\rvzhatzero \leftarrow \text{\texttt{Backward}} (\rvz_{t_i}; \rvs_\rvtheta)$ \Comment{Solve PF-ODE~(\ref{eq:pf-ode}) backward from $t=t_i$ to $t=0$}
    \State $\rvp \sim \mathcal N(\vzero, \mI)$
    \For{$j = 1, ..., M$}
    \State $\boldsymbol{\epsilon}_j \sim \mathcal{N}(\vzero, \mI)$
    \State $(\rvzhatzero, \rvp)$ = \texttt{Hamiltonian-Dynamics}($\rvzhatzero$, $\rvp$, $\boldsymbol{\epsilon}_j$; $\eta$, $\gamma$) \Comment{HMC updates for data consistency}
    \EndFor
    \State $\rvz_{t_{i-1}} \sim \mathcal{N}(\rvzhatzero, \sigma_{t_{i-1}}^2 \mI)$ \Comment{Proceed to the next noise level at time $t=t_{i-1}$}
\EndFor
\State \Return $\mathcal{D}(\rvz_{t_0})$ \Comment{Return the decoded \suppchange{video}}
\end{algorithmic}
\end{algorithm*}

\noindent The algorithm's main loop alternates between three key steps: (1) solving the PF-ODE backward from \( t = t_i \) to \( t = 0 \) (line 3), (2) performing multi-step MCMC updates (lines 4–8), and (3) advancing to the next noise level (line 9). We will discuss each step in detail.

\paragraph{Solving PF-ODE backward from $\bm{t=t_{i}}$ to $\bm{t=0}$}
The probability flow ordinary differential equation (PF-ODE) \citep{karras2022elucidating} of the diffusion model, given by \Eqref{eq:pf-ode}, governs the continuous increase or reduction of noise in the image when moving forward or backward in time. Here, \( \dot{\sigma}_t \) denotes the time derivative of \( \sigma_t \), and \( \nabla_{\rvz_t} \log p(\rvz_t; \sigma_t) \) represents the time-dependent score function \citep{song2019generative, song2021scorebased}.
\begin{equation}
\label{eq:pf-ode}
    \mathrm{d}\rvz_t = -\dot\sigma_t\sigma_t\nabla_{\rvz_t}\log p(\rvz_t;\sigma_t)\,\mathrm{d}t,
\end{equation}

Our goal is to solve the probability flow ODE (PF-ODE), as defined in \Eqref{eq:pf-ode}, backward from \( t = t_i \) to \( t = 0 \), given the intermediate state \( \rvz_{t_i} \) and the pretrained latent score function \( \rvs_\rvtheta(\rvz; \sigma) \approx \nabla_\rvz \log p(\rvz; \sigma) \). Any ODE solver, such as Euler’s method or the fourth-order Runge-Kutta method (RK4)~\citep{butcher2008numerical}, can be used to solve this problem. Following previous conventions~\citep{zhang2024improving}, we adopt a few-step Euler method for solving it efficiently.

\paragraph{\suppchange{Score from the noise-prediction network}}
\suppchange{The latent diffusion models are trained with the DDPM loss in \Eqref{eq:idm-loss}, whereas DAPS operates on the variance-exploding process $\rvz_t = \rvz_0 + \sigma_t\boldsymbol{\epsilon}$. Since $\sqrt{\bar\alpha_k}\,\rvz_0 + \sqrt{1-\bar\alpha_k}\,\boldsymbol{\epsilon} = \sqrt{\bar\alpha_k}\,(\rvz_0 + \sigma\boldsymbol{\epsilon})$ with $\sigma = \sqrt{(1-\bar\alpha_k)/\bar\alpha_k}$, the denoiser and the score at noise level $\sigma$ are
\begin{equation*}
D_\rvtheta(\rvz;\sigma) = \rvz - \sigma\,\boldsymbol{\epsilon}_\rvtheta\Big(\frac{\rvz}{\sqrt{1+\sigma^2}},\, k(\sigma)\Big), \qquad \rvs_\rvtheta(\rvz;\sigma) = \frac{D_\rvtheta(\rvz;\sigma) - \rvz}{\sigma^2},
\end{equation*}
where $k(\sigma)$ is the (continuous) DDPM timestep at which $\sqrt{(1-\bar\alpha_k)/\bar\alpha_k} = \sigma$~\citep{karras2022elucidating}. The linear schedule with $K=1000$ steps reaches $\sigma \approx 150$, which covers all noise levels $\sigma_t \le T = 100$ used by DAPS.}

\paragraph{Multi-step MCMC updates}
Any MCMC samplers can be used, such as Langevin Monte Carlo (LMC) and Hamiltonian Monte Carlo (HMC). For example, the LMC update with step size $\eta$ is
\begin{equation}
\label{eq:langevin-latent}
    \rvz_0^{+} = \rvz_0 + \eta \nabla_{\rvz_0} \log p(\rvy \mid \mathcal D(\rvz_0)) \nonumber + \eta \nabla_{\rvz_0} \log p(\rvz_0 \mid \rvz_t) + \sqrt{2\eta} \bm{\epsilon}.
\end{equation}
Note that the first gradient term can be computed with (\ref{eq:latent-forward-model}).
The second gradient term, on the other hand, can be calculated by
\begin{equation}
    \nabla_{\rvz_0}\log p(\rvz_0\mid \rvz_t) = \nabla_{\rvz_0} \log p(\rvz_t \mid \rvz_0) + \nabla_{\rvz_0}\log p(\rvz_0)\nonumber\approx \nabla_{\rvz_0}\log p(\rvz_t \mid \rvz_0) + \vs_\vtheta(\rvz_0, t_{\min}).\label{eq:score_estimate}
\end{equation}
This approximation holds for \( t_{\min} \approx 0 \), assuming that \( \rvz_0 \) lies close to the clean latent manifold~\citep{song2019generative}. To improve both convergence speed and approximation accuracy, the MCMC samplers are initialized with the solutions obtained from the previous PF-ODE step, leveraging its outputs as a warm start.

Note that during MCMC updates, the decoder \( \mathcal{D} \) needs to be evaluated multiple times in the backward pass. To accelerate this process, we adopt Hamiltonian Monte Carlo (HMC), which typically requires fewer steps for convergence, thereby speeding up the algorithm. For each multi-step MCMC update, we introduce an additional momentum variable \( \rvp \), initialized as \( \mathcal{N}(\vzero, \mI) \). The \texttt{Hamiltonian-Dynamics}(\( \rvz_0, \rvp, \boldsymbol{\epsilon}; \eta, \gamma \)) update with step size \( \eta \) and damping factor \( \gamma \) is given by:  
\begin{align}\label{eq:hmc-latent}
    \rvp^{+} &= (1-\gamma\eta)\cdot \rvp + \eta \nabla_{\rvz_0} \log p(\rvz_0 \mid \rvz_t) + \sqrt{2\gamma\eta} \bm{\epsilon}\\
    \rvz_0^{+} & = \rvz_0+\eta \rvp^+
\end{align}

\paragraph{Proceeding to next noise level}
According to Proposition 1 in~\citep{zhang2024improving}, one can obtain a sample $\rvz_{t_{i-1}}\sim p(\rvz_{t_{i-1}}\mid \rvy)$ by simply adding Gaussian noise from a sample $\hat \rvz_0\sim p(\rvz_0\mid \rvz_{t_{i}}, \rvy)$, given $\rvz_{t_{i}}\sim p(\rvz_{t_{i}}\mid \rvy)$ from last step. Thus we solve the target posterior sampling by gradually sampling from the time-marginal posterior of diffusion trajectory. The full hyperparameters and running cost of \ours~are summarized in Tab.~\ref{tab:app-step-hyper}. The HMC-related parameters are searched on a \suppchange{held-out} validation \suppchange{set of} 3 videos that are \suppchange{disjoint from the test} videos.

\begin{table}
    \centering
    \caption{\textbf{Summary of \ours~for black hole video reconstruction and dynamic MRI.} We provide and group the hyperparameters of \Algref{alg}. HMC related parameters are tuned on $3$ \suppchange{held-out} validation videos. The run time and memory are tested using 1 NVIDIA A100-SXM4-80GB GPU.}
    \label{tab:app-step-hyper}
    \resizebox{0.7\linewidth}{!}{
    \begin{tabular}{l!{\vrule}cc}
        \toprule
         & Black hole video reconstruction & Dynamic MRI \\
        \midrule
        \textbf{PF-ODE Related} & \\
        number of steps $N_{\text{ode}}$ & 20 & 20 \\
        scheduler $\sigma_t$ & $t$ &  $t$ \\
        \midrule
        \textbf{HMC Related} & \\
        number of steps $M$ & 60 &  53\\
        scaling factor $1-\gamma\eta$ & 0.00 & 0.83 \\
        step size square $\eta^2$ & 1.2e-5 & 1.2e-3 \\
        observation noise level $\sigma_{\rvy}$ & \suppchange{0.021} & \suppchange{0.10} \\
        \midrule
        \textbf{Decoupled Annealing Related} &\\
        number of steps $N$  & 25 & 20 \\
        final time $T$ & 100 & 100 \\
        discretization time $\{t_i\}, i=1,\cdots, N$ & $ \left(\tfrac{N - i}{N}\cdot T^\frac{1}{7}\right)^7$ & $\left(\tfrac{N - i}{N}\cdot T^\frac{1}{7}\right)^7$ \\ \midrule
        \textbf{Inference Related} & \\
        decoder NFE $N_\text{dec}$ & 1500 & 1060\\
        diffusion model NFE $N_\text{dm}$ & 500 & 400\\
        time (s) per sample & 645 & 332 \\
        memory (GB) & 48 & 21 \\
        \bottomrule
    \end{tabular}
    }
\end{table}

\newpage
\section{Experimental details}
\label{sec:app-exp-details}

\subsection{Black hole video reconstruction} 
\label{sec:bh-exp-details}
We introduce the black hole video reconstruction problem in more detail. The goal is to reconstruct a video $\rvx_0 \in \R^{n_f\times n_h \times n_w}$ of a rapidly moving black hole. 
Each measurement, or \textit{visibility}, is given by correlating the measurements from a pair of telescopes to sample a particular spatial Fourier frequency of the source with very long baseline interferometry (VLBI)~\citep{van1934wahrscheinliche, zernike1938concept}. In VLBI, the cross-correlation of the recorded scalar electric fields at two telescopes, known as the ideal \textit{visibility}, is related to the ideal source video $\rvx_0$ through a 2D Fourier transform, as given by the van Cittert-Zernike theorem \citep{van1934wahrscheinliche, zernike1938concept}.
Specifically, the ideal visibility of the $j$-th frame of the target video is 
\begin{equation}
    \rvI_{\{a,b\}}^{[j]}(\rvx_0) := \int_{\rho}\int_{\delta} \exp\left(-i2\pi\left(u_{\{a,b\}}^{[j]}\rho + v_{\{a,b\}}^{[j]}\delta\right)\right) \rvx_0^{[j]}(\rho,\delta) \mathrm{d}\rho \mathrm{d}\delta \in \mathbb{C},
\end{equation}
where $(\rho, \delta)$ denotes the angular coordinates of the source video frame, and $\left(u_{\{a,b\}}^{[j]}, v_{\{a,b\}}^{[j]}\right)$ is the dimensionless baseline vector between two telescopes $\{a, b\}$, orthogonal to the source direction.

Due to atmospheric turbulence and instrumental calibration errors, the observed visibility is corrupted by gain error, phase error, and additive Gaussian thermal noise \citep{m87paperiii, sun2024provable}:
\begin{equation}
    \rvV_{\{a,b\}}^{[j]} := g_a^{[j]} g_b^{[j]} \exp\left(-i\left(\phi_a^{[j]} - \phi_b^{[j]}\right)\right) \rvI_{\{a,b\}}^{[j]}\left(\rvx_0\right) + \rvn_{\{a,b\}}^{[j]} \in \mathbb{C}.
\end{equation}
where gain errors are denoted by $g_a^{[j]}, g_b^{[j]}$, phase errors are denoted by $\phi_a^{[j]}, \phi_b^{[j]}$, and thermal noise is denoted by $\rvn_{\{a,b\}}^{[j]}$.
While the phase of the observed visibility cannot be directly used due to phase errors, the product of three visibilities among any combination of three telescopes, known as the \textit{bispectrum}, can be computed to retain useful information. 
Specifically, the phase of the bispectrum, termed the \textit{closure phase}, effectively cancels out the phase errors in the observed visibilities.
Similarly, a strategy can be employed to cancel out amplitude gain errors and extract information from the visibility amplitude \citep{Blackburn_2020}.
Formally, these quantities are defined as
\begin{equation}
    \begin{aligned}
        \rvy_{\text{cp}, \{a,b,c\}}^{[j]} &:= \angle (\rvV_{\{a,b\}}^{[j]}\rvV_{\{b,c\}}^{[j]} \rvV_{\{a,c\}}^{[j]}) \in \R,\\
        \rvy_{\text{logca},\{a,b,c,d\}}^{[j]} &:= \log \left(\frac{\left|\rvV_{\{a,b\}}^{[j]}\right| \left|\rvV_{\{c,d\}}^{[j]}\right|}{\left|\rvV_{\{a,c\}}^{[j]}\right| \left|\rvV_{\{b,d\}}^{[j]}\right|}\right) \in \R.
    \end{aligned}
\end{equation}
Here, $\angle(\cdot)$ denotes the complex angle, and $|\cdot|$ computes the complex amplitude.
For a total of $P$ telescopes, the number of closure phase measurements $\rvy_{\text{cp}, \{a,b,c\}}^{[j]}$ is $\frac{(P-1)(P-2)}{2}$, and the number of log closure amplitude measurements $\rvy_{\text{logca},\{a,b,c,d\}}^{[j]}$ is $\frac{P(P-3)}{2}$, after accounting for redundancy. Let $D^{[j]}_\text{cp}$ and $D^{[j]}_\text{logca}$ denote the dimension of $\rvy_\text{cp}^{[j]}$ and $\rvy_\text{logca}^{[j]}$. Since closure quantities are nonlinear transformations of the visibilities, the black hole video reconstruction problem is non-convex. 

To aggregate data over different measurement times and telescope combinations, the forward model of black hole video reconstruction for the $j$-th frame can be expressed as
\begin{equation}
    \rvy^{[j]} := \left[\mathcal{A}_\text{cp}^{[j]}(\rvx_0), \mathcal{A}_\text{logca}^{[j]}(\rvx_0), \mathcal{A}_\text{flux}^{[j]}(\rvx_0)\right] := \left[\rvy_\text{cp}^{[j]}, \rvy_\text{logca}^{[j]}, \rvy_\text{flux}^{[j]}\right],
\end{equation}
where \(\rvy_\text{cp}^{[j]} = \left[\rvy_{cp, \{a,b,c\}}^{[j]}\right]\) is the set of all closure phase measurements and \(\rvy_\text{logca}^{[j]} = \left[\rvy_{logca, \{a,b,c,d\}}^{[j]}\right]\) is the set of all log closure amplitude measurements for $j$-th frame.
The total flux at the $j$-th frame, representing the DC component of the Fourier transform, is given by
\begin{equation}
    \rvy_\text{flux}^{[j]} := \int_{\rho}\int_{\delta} \rvx_0^{[j]}(\rho,\delta) \mathrm{d}\rho \mathrm{d}\delta.
\end{equation}
The overall data consistency is an aggregation over all frames and typically expressed using the $\boldsymbol{\chi}^2$ statistics

\begin{align}\label{app:eq-chisquare}
    -\log p(\rvy\mid \rvx_0) &\propto  \underbrace{\sum_{j=1}^{n_{f}}\dfrac{1}{n_fD_{\text{cp}}^{[j]} \sigma_\text{cp}^2 }\left\| \mathcal A^{[j]}_\text{cp}(\rvx_0) - \rvy^{[j]}_\text{cp}\right\|^2}_{\boldsymbol{\chi}_{\text{cp}}^2} + \underbrace{\sum_{j=1}^{n_{f}}\dfrac{1}{n_fD_{\text{logca}}^{[j]} \sigma_\text{logca}^2 }\left\| \mathcal A^{[j]}_\text{logca}(\rvx_0) - \rvy^{[j]}_\text{logca}\right\|^2}_{\boldsymbol{\chi}_{\text{logca}}^2} \nonumber\\
    &+ \beta\underbrace{\sum_{j=1}^{n_{f}}\dfrac{1}{n_f \sigma_\text{flux}^2 }\left\| \mathcal A^{[j]}_\text{flux}(\rvx_0) - \rvy^{[j]}_\text{flux}\right\|^2}_{\boldsymbol{\chi}_{\text{flux}}^2}
\end{align}
where $\sigma_\text{cp}$, $\sigma_\text{logca}$, and $\sigma_\text{flux}$ are the estimated standard deviations of the measured closure phase, log closure amplitude, and flux, respectively, and $\beta$ is a hyperparameter that controls the strength of the flux regularization, which is empirically determined. To evaluate the data fitting, we introduce a unified $\bm{\tilde\chi}^2$ statistics
\begin{equation}
    \label{app:eq-chisquare-metrics}
\tilde{\boldsymbol{\chi}}^2 = \boldsymbol{\chi}^2 \cdot \mathds{1}\{\boldsymbol{\chi}^2 \geq 1\} + \frac{1}{\boldsymbol{\chi}^2} \cdot \mathds{1}\{\boldsymbol{\chi}^2 < 1\}.
\end{equation}
The $\tilde{\boldsymbol{\chi}}^2$ is no less than $1$ and closer to $1$, indicating better measurement data fit. In our experiments we use an average $(\tilde{\boldsymbol{\chi}}^2_{\text{cp}} + \tilde{\boldsymbol{\chi}}^2_\text{logca})/2$ for the data misfit metrics for evaluation.

\begin{figure}
    \centering
    \includegraphics[width=\linewidth]{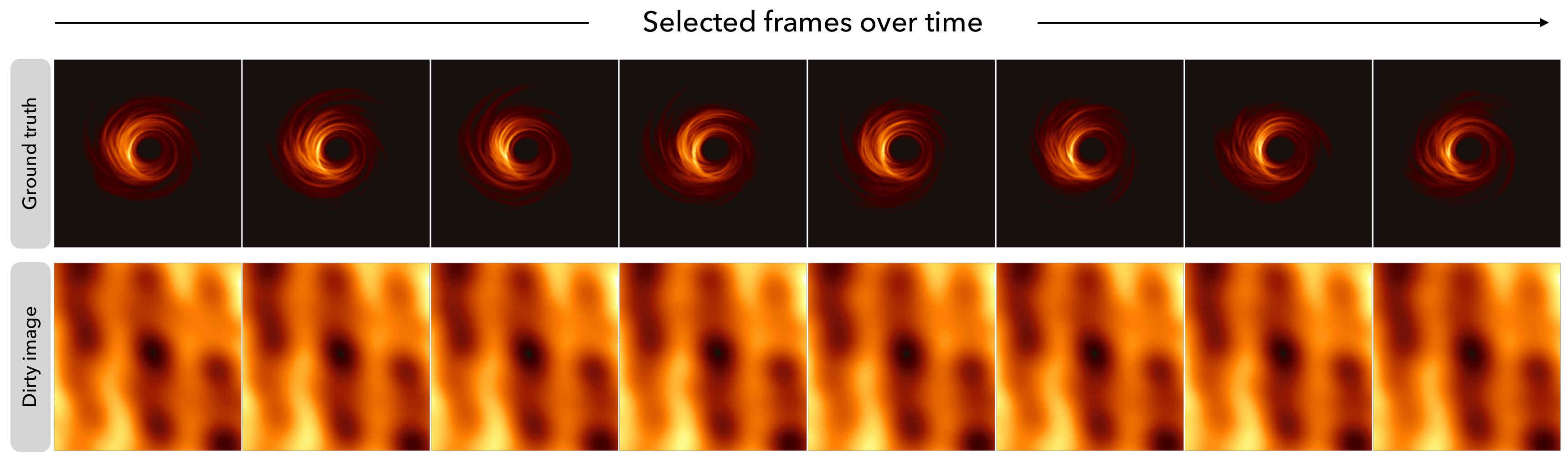}    
    \caption{\textbf{The dirty images from the ideal visibilities}. We use the standard implementation in \suppchange{the \texttt{eht-imaging} library~\citep{chael2016high, chael2018interferometric}} to get dirty images for each selected frame.}
    \label{fig:app-dirty-image}
\end{figure}

Our experiments are based on the simulation of observing the Sagittarius A$^\ast$ black hole with the EHT 2017 array of eight radio telescopes over an observation period of around 100 minutes\suppchange{, during which seven of them observe the source}.
We refer the readers to Fig. 5 of~\citep{levis2021fluid} for a visualization of the measurement patterns in Fourier space over time.
To show the difficulty of this black hole video reconstruction problem, we visualize the dirty video frames obtained by applying the inverse Fourier transform to the ideal visibilities, assuming no measurement errors, in \Figref{fig:app-dirty-image}. One can see that substantial spatiotemporal information is lost during the measurement process, so obtaining high-quality reconstructions relies on the effectiveness of incorporating prior information in the reconstruction process.

\subsection{Dynamic MRI}
\label{sec:mri-exp-details}
MRI is an important imaging technique for clinical diagnosis and biomedical research, where the objective is to recover a video $\rvx_0 \in \mathbb{C}^{n_f\times n_h \times n_w}$ of the heart from the subsampled Fourier space (a.k.a $k$-space) measurements $\rvy$. Despite its many advantages, MRI is known to be slow because of the physical limitations of the data acquisition in $k$-space.
This leads to low patient throughput and sensitivity to patient's motion~\citep{wang2024cmrxrecon}.
To accelerate the scan speed, instead of fully sampling $k$-space, the compressed subsampling MRI (CS-MRI) technique subsamples $k$-space with masks $\{\rvm^{[j]}\}_{j=1}^{n_f}$. Mathematically, this can be formulated as
\begin{align}
    \rvy^{[j]} = \rvm^{[j]} \odot \mathcal{F}\left(\rvx_0^{[j]}\right) + \rvn^{[j]} \in \mathbb{C}^n \quad \text{for } j = 1, ..., n_f,
\end{align}
where $\rvm^{[j]} \in \{0, 1\}^{n_h \times n_w}$ is the subsampling mask for the $j$-th frame, $\odot$ denotes element-wise multiplication, $\mathcal{F}$ is the Fourier transform, and $\rvn^{[j]}$ is the measurement noise. 
In our experiments, we used subsampling masks with an equi-spaced pattern (similar to those visualized in~\cite{wang2024cmrxrecon}) of both $6\times$ acceleration with 24 auto-calibration signal (ACS) lines (Tab.~\ref{tab:app-dynamic-mri}) and $8\times$ acceleration with 12 ACS lines (Tab.~\ref{tab:main-table}). For dynamic MRI, we use the Gaussian likelihood function
\begin{equation}
    -\log p(\rvy|\rvx_0) \propto \|\mathcal A(\rvx_0)-\rvy\|^2_2.
\end{equation}

\begin{figure}[t!]
    \centering
    \includegraphics[width=0.9\linewidth]{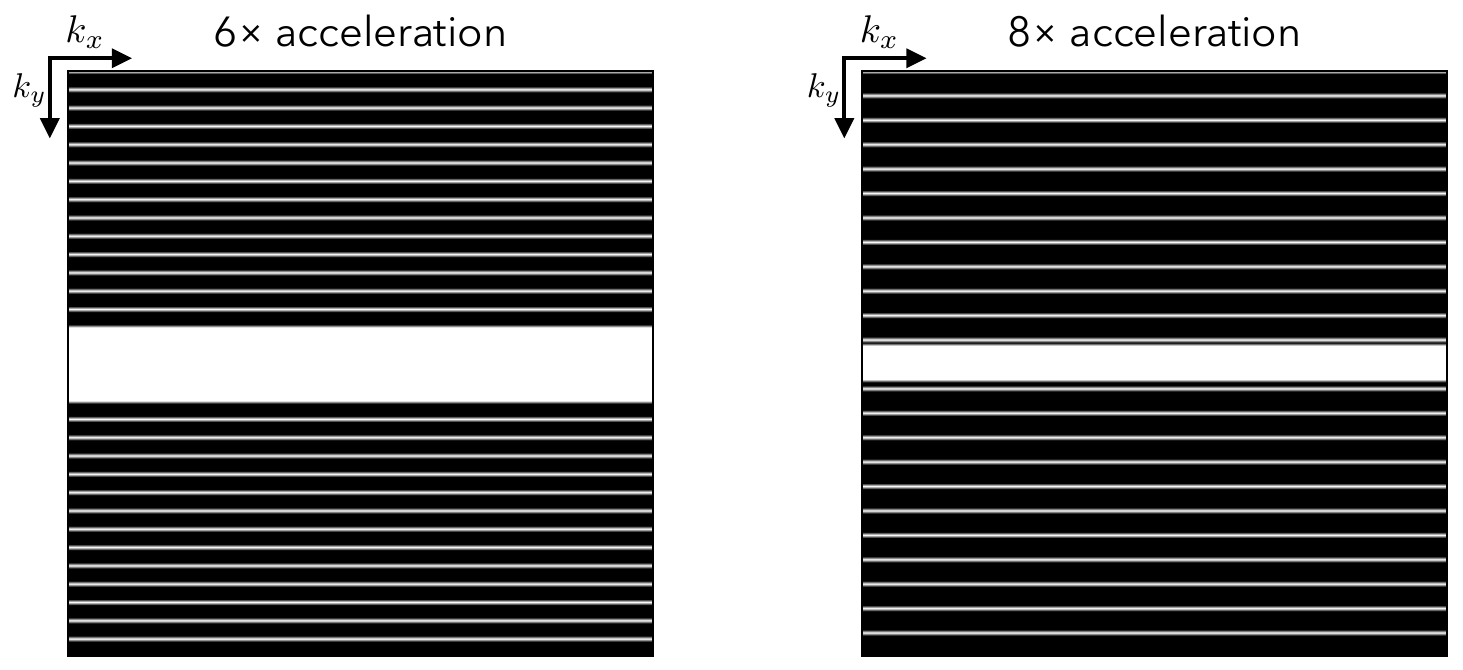}
    \caption{\textbf{Subsampling masks of $6\times$ (\underline{left}) and $8\times$ (\underline{right}) accelerations used in dynamic MRI experiments.} The white areas in the center indicate the auto-calibration (ACS) signals. The horizontal and vertical directions are the frequency ($k_x$) and phase ($k_y$) encoding directions, respectively. The same mask is applied to the sampling of each individual frame of all videos.}
    \label{fig:mri-mask}
\end{figure}

Fig.~\ref{fig:mri-mask} visualizes the subsampling masks used in our experiments, where $k_x, k_y$ indicate the frequency encoding and phase encoding directions, respectively.
The same mask is applied to the sampling of each frame of all videos.

\subsection{Baseline implementations}
\label{sec:baseline-implementation}
In this section, we explain the detailed implementation of our baseline methods. We follow the same grouping in Sec.~\ref{sec:baselines}.

\paragraph{Group 1: simple heuristics}
We implement \emph{batch independent sampling} (BIS) and \emph{batch consistent sampling} (BCS) following~\citep{kwon2025solving,kwon2024visionxlhighdefinitionvideo}. BIS is implemented to reconstruct each frame independently, whereas BCS promotes temporal consistency by using identical ("batch-consistent") noise across the temporal dimension. To ensure a fair comparison, we adapt \Algref{alg} by initializing the noise variables $\rvz_{t_{N}}$ (line 1) and $\rvz_{t_{i-1}}$ (line 9) with batch-consistent noise.

\paragraph{Group 2: noise warping}
We follow~\citep{deng2025infiniteresolution,chang2024how} in performing noise warping based on optical flow estimated from measurements. As described in Sec.~\ref{sec:baselines}, we derive optical flow using ground-truth black hole videos and inverse Fourier-transformed dynamic MRI videos with a pretrained model~\citep{teed2020raftrecurrentallpairsfield}. Since we utilize a latent diffusion model, we further downsample the optical flow via interpolation to match the latent noise dimension. We adapt \Algref{alg} accordingly by replacing the original \emph{i.i.d.} noise with the warped noise while keeping other components unchanged. We implement \emph{Bilinear}, \emph{Bicubic}, and \emph{Nearest} warping strategies using their corresponding interpolation methods, and follow the implementation from \url{https://github.com/yitongdeng-projects/infinite_resolution_integral_noise_warping_code} for $\int$-noise. Lastly, since~\citep{daras2024warped} does not provide publicly available code, we implement GP-Warp following Equation (2) from their paper.

\paragraph{\suppchange{Group 3: Wan2.1 + LoRA}}
\suppchange{We use the base Wan2.1-T2V-1.3B model~\citep{wan2025wan} rather than the few-step distilled Wan models that LVTINO and AVIS use as priors, because those are trained to denoise only at a few fixed noise levels, whereas DAPS evaluates the denoiser along the whole annealing schedule.

\textbf{Data and input format.} We train on the same 648 training videos as \ours~(video only), with the same normalization to $[-1,1]$ and the same augmentation (random flips and zooms), and without any image data. Each grayscale frame is replicated to three channels. The causal 3D VAE of Wan2.1 compresses space by $8\times$ and time by $4\times$ and requires $1+4k$ frames: the first latent frame encodes one frame and each subsequent latent frame encodes four. We therefore prepend a copy of the first frame to each 64-frame training clip (a random window of 64 consecutive frames, as for \ours; Appendix~\ref{sec:app-training-details}), which gives 65 frames and a latent of size $16\times17\times32\times32$, in which each of the last 16 latent frames covers four consecutive frames of the original video. Latents are normalized per channel with the statistics provided with Wan2.1. After decoding, we drop the duplicated first frame and average the three channels, so the forward model receives a $64\times256\times256$ video as for all other methods.

\textbf{Conditioning.} The model is text-conditioned. We use one fixed prompt that describes the domain, ``A grayscale video of a black hole: a bright ring of hot glowing gas around a dark central shadow, slowly swirling and flickering, on a black background.'', encode it once with the umT5 text encoder of Wan2.1, and use this embedding for both fine-tuning and sampling. We do not use classifier-free guidance, which would distort the learned distribution, so the model represents a single prior over black hole videos. We do not use per-video captions, since properties of the test videos are unknown at inference time.

\textbf{LoRA fine-tuning.} We insert rank-64 LoRA adapters ($\alpha=64$) into the query, key, value, and output projections of the self-attention layers and into both linear layers of the feed-forward networks of all 30 transformer blocks, which gives 63.9M trainable parameters (about half of the 138.0M of the spatiotemporal UNet of \ours); all pretrained weights, the VAE, and the text encoder stay frozen. We fine-tune with the flow-matching loss used to pretrain Wan2.1~\citep{lipman2023flow}, sampling $\tau$ uniformly: with $\mathbf{u}_\tau=(1-\tau)\rvz_0+\tau\boldsymbol{\epsilon}$, $\boldsymbol{\epsilon}\sim\mathcal N(\bm 0,\bm I)$, and $\tau\sim\mathcal U(0,1)$, we minimize $\mathbb{E}\,\|\mathbf{v}_{\bm\theta}(\mathbf{u}_\tau,\tau)-(\boldsymbol{\epsilon}-\rvz_0)\|_2^2$. We use AdamW with a learning rate of $10^{-4}$, a batch size of 8, and bf16 mixed precision, and train for 500 epochs, the same as the spatiotemporal UNet of \ours~(Tab.~\ref{tab:app-vdm-training}), which takes 22 hours on 4 A100 GPUs.

\textbf{Use in DAPS.} DAPS operates on the normalized latents, with $\rvz_t=\rvz_0+\sigma_t\boldsymbol{\epsilon}$ (\Algref{alg}). Since $\rvz_t/(1+\sigma_t)=(1-\tau)\rvz_0+\tau\boldsymbol{\epsilon}$ with $\tau=\sigma_t/(1+\sigma_t)$, the denoiser required by DAPS is
\begin{equation*}
D_{\bm\theta}(\rvz_t;\sigma_t)=\mathbf{u}_\tau-\tau\,\mathbf{v}_{\bm\theta}(\mathbf{u}_\tau,\tau),\qquad \mathbf{u}_\tau=\frac{\rvz_t}{1+\sigma_t},\quad \tau=\frac{\sigma_t}{1+\sigma_t},
\end{equation*}
so all noise levels $\sigma_t\le100$ used by DAPS correspond to $\tau\le0.990$; the transformer receives $1000\tau$ as its timestep input. We use the same sampler settings as for \ours~(Tab.~\ref{tab:app-step-hyper}), including the initialization from Gaussian noise, and retune only the HMC step size on the same three validation videos ($\eta^2=6\times10^{-6}$). Each HMC step backpropagates through the frozen 3D VAE decoder; the transformer and the decoder run in bf16. We evaluate on the same measurements of the 10 test videos as all other methods and measure sampling time and memory on the same type of GPU as in Tab.~\ref{tab:app-step-hyper}.

\textbf{Cost and comparability.} The Wan2.1 denoiser has 1.4B parameters, about 10$\times$ more than the 138.0M of the spatiotemporal UNet of \ours, and sampling takes 2542~s and 64~GB per video, versus 645~s and 48~GB for \ours~(Tab.~\ref{tab:app-step-hyper}). Besides pretraining, the two priors differ in other respects: Wan2.1 uses a causal 3D VAE that compresses time by $4\times$, is adapted with LoRA rather than fully fine-tuned, and is text-conditioned. The comparison therefore does not isolate the effect of pretraining on natural videos.}

\begin{table}[b!]
    \centering
    \caption{\textbf{Summary of the training of spatiotemporal diffusion prior}. We provide and group the hyperparameters according to each component in the model. \suppchange{All stages use Adam and keep an EMA of the weights. Batch sizes of the IDM and of joint fine-tuning count (pseudo-)videos of $n_f$ frames.} The model is trained with 1 NVIDIA A100-SXM4-80GB GPU.}
    \label{tab:app-vdm-training}
    \resizebox{0.7\linewidth}{!}{
    \begin{tabular}{l!{\vrule}cc}
        \toprule
        Hyper-parameters & Black hole video reconstruction & Dynamic MRI \\
        \midrule
        \textbf{Dataset Related} &\\
        frames $n_f$ & 64 & 12\\
        \suppchange{frames per training video} & \suppchange{1000} & \suppchange{12}\\
        resolution $n_h\times n_w$ & 256$\times$256 & 192$\times$192\\
        $N_{\text{image}}$ & 50000 & 39888\\
        $N_{\text{video}}$ & 648 & 3324\\
        \suppchange{augmentation} & \suppchange{flips, zooms} & \suppchange{flips}\\
        \textbf{VAE Training Related} & \\
        latent channels & 1  & 2 \\
        block channels & [64, 128, 256, 256] & [256, 512, 512] \\
        down sampling factor & 8 & 4\\
        batch size & \suppchange{64} & 16\\
        \suppchange{learning rate} & \suppchange{2e-4} & \suppchange{2e-4}\\
        \suppchange{EMA decay} & \suppchange{0.99} & \suppchange{0.99}\\
        epochs & 25 & 10\\
        $\beta_\text{KL}$ & \suppchange{1e-6} & \suppchange{1e-6} \\
        \midrule
        \textbf{IDM Training Related} & \\
        block channels & [128, 256, 512, 512] & [128, 256, 512, 512]\\
        \suppchange{noise schedule} & \multicolumn{2}{c}{\suppchange{linear, $\beta_k$ from $10^{-4}$ to $0.02$, $K=1000$}}\\
        batch size & \suppchange{4} & 16\\
        \suppchange{learning rate} & \suppchange{2e-4} & \suppchange{2e-4}\\
        \suppchange{EMA decay} & \suppchange{0.999} & \suppchange{0.999}\\
        epochs & 200 & 50\\
        \midrule
        \textbf{Joint Fine-tuning Related} & \\
        \suppchange{$p_{\text{img}}$} & \suppchange{0.2} & \suppchange{0.2}\\
        \suppchange{batch size} & \suppchange{4} & \suppchange{16}\\
        \suppchange{learning rate} & \suppchange{2e-4} & \suppchange{2e-4}\\
        \suppchange{EMA decay} & \suppchange{0.999} & \suppchange{0.999}\\
        epochs & 500 & 300 \\
        \midrule
        \textbf{Other Info} & \\
        VAE parameters & \suppchange{15.5M} & \suppchange{60.3M}\\
        \suppchange{spatiotemporal UNet} parameters & \suppchange{138.0M} & \suppchange{138.0M}\\
        VAE training time & 4.5h & 8.9h \\
        image diffusion model training time & 5.5h & 3.8h\\
         joint fine-tuning time & 13.7h & 22.8h\\
        \bottomrule
    \end{tabular}
    }
\end{table}

\section{Training details for spatiotemporal diffusion prior}
\label{sec:app-training-details}

In this section, we \suppchange{describe how we train the spatiotemporal diffusion priors for} black hole video reconstruction and dynamic MRI, and we summarize the training hyper-parameters in \Tabref{tab:app-vdm-training}. We define \( D_{\text{image}} \) and \( D_{\text{video}} \) as the image and video datasets, containing \( N_{\text{image}} \) and \( N_{\text{video}} \) data points, respectively. \suppchange{For black holes, \( D_{\text{image}} \) is an independent set of 50{,}000 simulated GRMHD images drawn i.i.d.\ from the same distribution as the video frames; it is separate from the 648 training videos and from the test and validation videos. Each training video is a simulation of 1{,}000 frames, and each training iteration uses a random window of $n_f = 64$ consecutive frames. For dynamic MRI, no separate image set from the same distribution is available, so \( D_{\text{image}} \) consists exactly of the individual frames of \( D_{\text{video}} \) ($39{,}888 = 3{,}324\times 12$), and no additional data are used. For both tasks, the 10 test videos and 3 validation videos were drawn i.i.d.\ and held out at the outset; they appear in no training data, neither as videos nor as frames of \( D_{\text{image}} \).}
For data augmentation, we apply random horizontal/vertical flipping \suppchange{for both tasks and, for black holes, random zoom-in-and-out with a scale factor in $[0.75, 1.25]$} to improve robustness and diversity in training.

\suppchange{\textbf{Optimization.} All stages use Adam with a learning rate of $2\times10^{-4}$ and keep an exponential moving average (EMA) of the weights, with a decay of 0.99 for the VAE and 0.999 for the diffusion models. Each stage starts from the EMA weights of the previous stage, and the EMA weights of the final models are used for sampling.}

We first train the compression functions, the encoder \( \mathcal{E} \) and decoder \( \mathcal{D} \), on an image dataset. The training objective consists of an L1 reconstruction loss combined with a KL divergence term scaled by a factor \( \beta_{KL} \)\suppchange{, as defined in \Eqref{eq:vae-loss} (\Secref{sec:vdm-prior}), where the L1 error is averaged over pixels and the KL divergence is summed over latent dimensions. The KL weight trades off two effects: a larger $\beta_\text{KL}$ degrades the VAE reconstructions, which limits the quality of every decoded sample, whereas a too small $\beta_\text{KL}$ leaves the latent space poorly regularized, which noticeably slows the convergence of diffusion training. We use $\beta_\text{KL} = 10^{-6}$ for both tasks, which keeps the reconstructions accurate (\Figref{fig:app-vae})}.
Next, we train the image diffusion UNet \suppchange{$\boldsymbol{\epsilon}_\rvtheta$ with the DDPM noise-prediction loss in \Eqref{eq:idm-loss}, using the linear schedule of~\citet{ho2020denoising}, in which $\beta_k$ increases from $10^{-4}$ to $0.02$ over $K=1000$ steps}.

\suppchange{\textbf{Spatiotemporal UNet.} We implement the spatiotemporal UNet with the spatiotemporal blocks of \texttt{diffusers}~\citep{von-platen-etal-2022-diffusers}, originally developed for SVD~\citep{blattmann2023stablevideodiffusionscaling}, without text or image conditioning. Each spatial module (a ResNet or transformer block) is followed by a temporal module (a temporal ResNet block with 1D convolutions over frames, or a temporal transformer block) that takes the output of the spatial module as input. The two outputs are blended as in \Secref{sec:vdm-prior} with $\alpha = 1 - \mathrm{sigmoid}(m)$, where $m$ is a learnable scalar initialized to $0.5$, so that $\alpha \approx 0.38$ at initialization (rather than $0$, as drawn in \Figref{fig:st-module}); for image inputs, $\alpha = 0$. In our implementation, the IDM of Step 1 is this network trained with the switch \texttt{OFF}, so only the spatial modules are trained in Step 1 and the temporal modules keep their random initialization until Step 3.}

After \suppchange{Step 1}, \suppchange{the temporal modules are switched on and} the model is fine-tuned jointly using both video and image datasets. We use the same loss as in \Eqref{eq:idm-loss}, by changing \( \rvx_0 \) to a video or a pseudo-video input. \suppchange{Each iteration takes a gradient step on a batch of videos and, with probability $p_\text{img} = 0.2$, an additional step on a batch of pseudo-videos of independent images.} Each frame is independently processed using the encoder $\mathcal E$, ensuring that spatial representations remain aligned while temporal consistency is learned through the added temporal modules.

\suppchange{\ours~(video only) uses the same VAE, architecture, and settings, but its spatiotemporal UNet is randomly initialized and trained on videos only ($p_\text{img} = 0$) for the same number of epochs.}

\section{\suppchange{Fr\'echet Video Distance}}
\label{sec:app-fvd}
\suppchange{Tab.~\ref{tab:app-fvd} reports the FVD~\citep{unterthiner2018towards} between the reconstructions and the test videos, computed with \url{https://github.com/JunyaoHu/common_metrics_on_video_quality}. We report FVD only as a relative comparison between methods and do not base our conclusions on it: with 10 videos per set the estimate is strongly biased, and its features come from a network trained on natural videos, which is not designed for black hole or cardiac data.}

\begin{table}[h]
    \suppchangeon
    \centering
    \small
    \caption{\suppchange{\textbf{FVD ($\downarrow$) over the 10 test videos.} Same methods and test videos as in Tabs.~\ref{tab:main-table} and~\ref{tab:app-dynamic-mri}; the best result is in bold and the second best is underlined.}}
    \label{tab:app-fvd}
    \begin{tabular}{l!{\vrule}ccc}
        \toprule
        \textbf{Methods} & \textbf{Black hole} & \textbf{MRI ($8\times$)} & \textbf{MRI ($6\times$)} \\
        \midrule
        BIS~\citep{kwon2025solving} & 1429.42 & 82.30 & 113.17 \\
        BCS~\citep{kwon2025solving} & 564.43 & 95.68 & 110.01 \\
        Bilinear~\citep{chang2024how} & 1335.95 & 87.90 & 114.37 \\
        Bicubic~\citep{chang2024how} & 1521.14 & 108.90 & 106.82 \\
        Nearest~\citep{chang2024how} & 1171.07 & 108.57 & 110.91 \\
        $\int$-noise~\citep{deng2025infiniteresolution} & 1253.81 & 89.59 & 111.92 \\
        GP-Warp~\citep{daras2024warped} & 1395.15 & 92.19 & 105.70 \\
        \midrule
        Wan2.1 + LoRA~\citep{wan2025wan,hu2022lora} & 256.43 & -- & -- \\
        \midrule
        \ours~(video only) & \underline{238.36} & \textbf{75.27} & \underline{101.83} \\
        \ours~(image-video joint) & \textbf{170.67} & \underline{78.60} & \textbf{98.15} \\
        \bottomrule
    \end{tabular}
\end{table}

\section{Licenses}
\label{app:license}
We list the licenses of all the assets we used in this paper:
\begin{itemize}
    \item Data
    \begin{itemize}
        \item GRMHD black hole data: Unknown
        \item CMRxRecon Challenge 2023~\cite{wang2024cmrxrecon}: MIT License
    \end{itemize}
    \item Code
    \begin{itemize}
     \item $\int$-noise~\citep{deng2025infiniteresolution}\footnote{$\int$-noise: \url{https://github.com/yitongdeng-projects/infinite_resolution_integral_noise_warping_code}}: MIT License  
        \item Diffusers~\citep{von-platen-etal-2022-diffusers}\footnote{Diffusers: \url{https://github.com/huggingface/diffusers}}: Apache-2.0 license
        \item \suppchange{Wan2.1~\citep{wan2025wan}\footnote{\suppchange{Wan2.1: \url{https://github.com/Wan-Video/Wan2.1}}}: Apache-2.0 license}

    \item Common metrics on video quality\footnote{Common metrics on video quality: \url{https://github.com/JunyaoHu/common_metrics_on_video_quality}}: None
    \end{itemize}
    
\end{itemize}

\newpage
\section{More results \& visualization}
\label{sec:app-more-res-visual}

\begin{table}[t!]
\scriptsize
\centering
\caption{\textbf{Additional results on dynamic MRI with 6$\times$ acceleration.} \suppchange{Same setup and metrics as for dynamic MRI ($8\times$) in Tab.~\ref{tab:main-table}, with the same prior.}}
\label{tab:app-dynamic-mri}
\resizebox{\linewidth}{!}{
\begin{tabular}{lccc!{\vrule}cc!{\vrule}c}
    \toprule
    \multicolumn{7}{c}{\textbf{Task: MRI ($6\times$)}}\\
    \midrule
    \textbf{Methods} & \textbf{PSNR ($\uparrow$)} & \textbf{SSIM ($\uparrow$)} & \textbf{LPIPS ($\downarrow$)} & \textbf{d-PSNR ($\uparrow$)} & \textbf{d-SSIM ($\uparrow$)} & \textbf{Data Misfit ($\downarrow$)} \\ \midrule
    BIS~\citep{kwon2025solving} & 39.47 (0.59) & 0.958 (0.007) & 0.086 (0.011) & 43.26 (1.23) & 0.962 (0.005) & 11.071 (0.740) \\
    BCS~\citep{kwon2025solving} & 40.69 (0.57) & 0.959 (0.006) & 0.081 (0.012) & 44.73 (1.31) & 0.974 (0.004) & 11.085 (0.720) \\
    Bilinear~\citep{chang2024how} & \underline{40.85} (0.57) & 0.960 (0.006) & 0.080 (0.012) & 44.84 (1.28) & 0.975 (0.004) & 11.038 (0.735) \\
    Bicubic~\citep{chang2024how} & 40.71 (0.67) & 0.959 (0.007) & 0.079 (0.012) & 44.74 (1.38) & 0.974 (0.005) & 11.068 (0.755) \\
    Nearest~\citep{chang2024how} & 40.37 (0.56) & 0.960 (0.007) & 0.080 (0.012) & 44.81 (1.41) & 0.974 (0.005) & 11.050 (0.739) \\
    $\int$-noise~\citep{deng2025infiniteresolution} & 40.09 (0.50) & 0.960 (0.006) & 0.082 (0.012) & 44.77 (1.34) & 0.974 (0.005) & 11.059 (0.731) \\
    GP-Warp~\citep{daras2024warped} & 39.50 (0.48) & 0.959 (0.007) & 0.080 (0.012) & 44.53 (1.33) & 0.973 (0.005) & 11.070 (0.727) \\
    \midrule
    \ours~(video only) & 40.76 (0.43) & \underline{0.967} (0.005) & \underline{0.077} (0.012) & \underline{46.38} (1.82) & \underline{0.981} (0.005) & \textbf{10.788} (0.713) \\
    \ours~(image-video joint) & \textbf{41.39} (0.52) & \textbf{0.969} (0.005) & \textbf{0.076} (0.012) & \textbf{46.61} (1.72) & \textbf{0.982} (0.004) & \underline{10.808} (0.723) \\
    \bottomrule
\end{tabular}}
\end{table}

\begin{figure}[t]
    \centering
    \includegraphics[width=\linewidth]{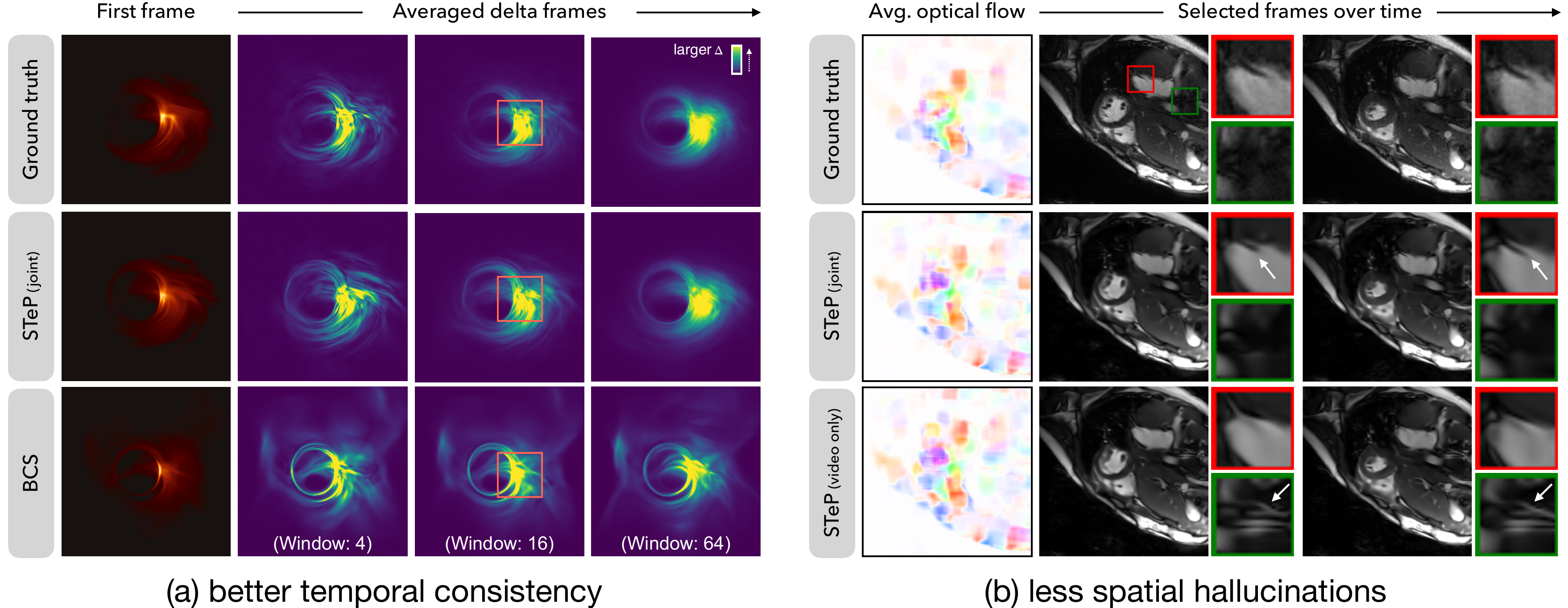}
    \caption{\textbf{Detailed comparison on black hole video reconstruction and dynamic MRI.}
    \suppchange{(a) Black hole:} We compare \ours~(joint) and the BCS baseline~\citep{kwon2025solving} by visualizing the averaged delta frames (difference images) over an expanding window. The delta frames given by \ours~(joint) better align with the ground truth, indicating better temporal consistency. \suppchange{(b) Dynamic MRI:} We also compare the spatial fidelity between \ours~(joint) and its variant \ours~(video only). Trained on both images and videos, \ours~(joint) provides reconstructions with fewer spatial hallucinations \suppchange{(arrows)} compared to \ours~(video only).}
    \label{fig:detailed-comparison}
\end{figure}

\begin{figure}[t]
    \centering
    \includegraphics[width=\linewidth]{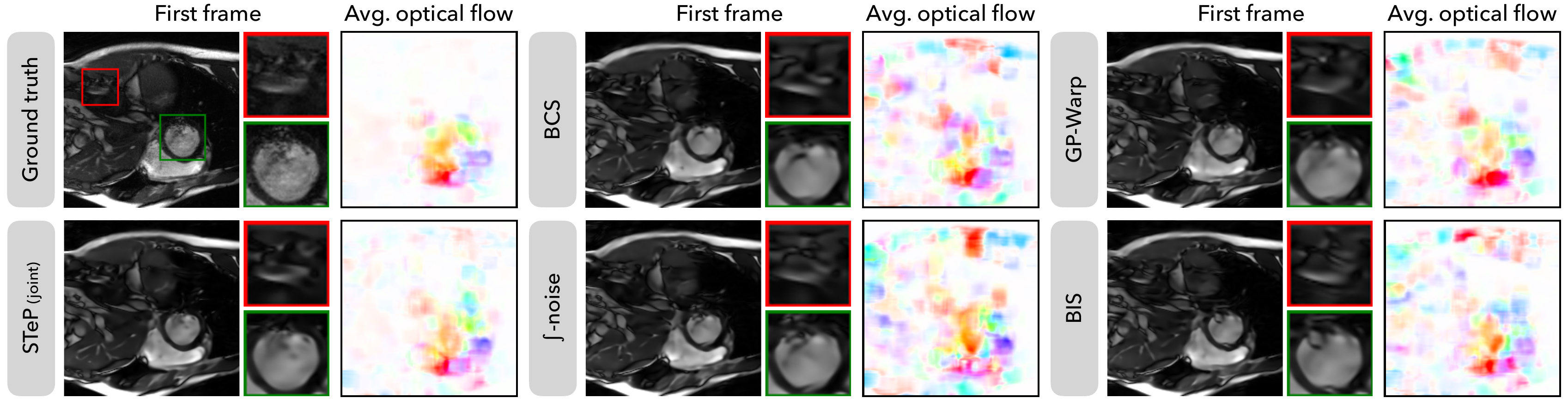}
    \caption{\textbf{Visual examples of \ours~(bottom left) and baselines for dynamic MRI \suppchange{($8\times$)}.} We visualize \suppchange{the first} frame along with two zoomed-in regions for each method to better illustrate spatial fidelity. \ours~provides reconstructions with fewer structural artifacts and temporal fluctuations, as indicated by its averaged optical flow aligning more closely with the ground truth.}
    \vspace{-0.12in}
    \label{fig:dynamic-mri}
\end{figure}

\begin{figure}[t]
    \centering
    \includegraphics[width=\linewidth]{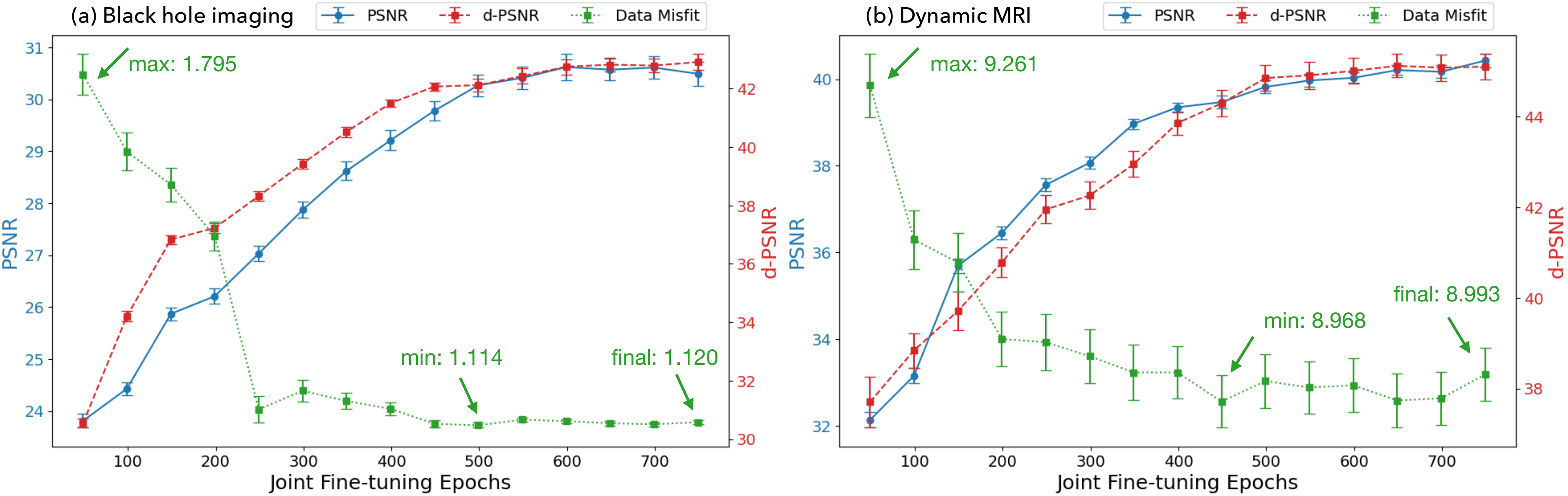} 
    \caption{\textbf{Consistent improvement in image-video joint fine-tuning.} We evaluate intermediate checkpoints of (a) black hole video reconstruction and (b) dynamic MRI (8$\times$) \suppchange{from training runs in which joint fine-tuning is continued to 750 epochs for this analysis. For black holes, the epoch-500 checkpoint of this run is the prior used in Tab.~\ref{tab:main-table}. For dynamic MRI, the curve comes from a separate run with the same settings, so its values need not match the MRI results in Tab.~\ref{tab:main-table}, which use a prior trained for 300 epochs (Tab.~\ref{tab:app-vdm-training})}.
    Spatial similarity (measured by PSNR), temporal consistency (measured by d-PSNR) and measurement data fit (measured by data misfit) all improve \suppchange{over the first few hundred epochs} and then plateau.}
    \vspace{-8pt}
    \label{fig:ablation-posterior}
\end{figure}

\paragraph{\suppchange{Paired differences.}}
\suppchange{Tab.~\ref{tab:app-paired} compares \ours~(image-video joint) with the strongest IDM-based baseline (BCS) and with \ours~(video only) on each of the 10 black hole test videos, using the same samples as in Tab.~\ref{tab:main-table}.}

\begin{table}[h]
    \suppchangeon
    \centering
    \scriptsize
    \caption{\suppchange{\textbf{Paired differences on black hole video reconstruction.} For each test video, we subtract the metric of the baseline from that of \ours~(image-video joint), computed as in Tab.~\ref{tab:main-table}. We report the mean and standard deviation (in parentheses) of the difference over the 10 test videos, and in brackets the number of videos on which \ours~(image-video joint) is better (higher PSNR, SSIM, d-PSNR, and d-SSIM; lower LPIPS and data misfit).}}
    \label{tab:app-paired}
    \resizebox{\linewidth}{!}{
    \begin{tabular}{l!{\vrule}ccc!{\vrule}cc!{\vrule}c}
        \toprule
        \textbf{Difference} & \textbf{PSNR} & \textbf{SSIM} & \textbf{LPIPS} & \textbf{d-PSNR} & \textbf{d-SSIM} & \textbf{Data Misfit} \\
        \midrule
        \ours~(joint) $-$ BCS & $+$2.62 (1.02) [10/10] & $+$0.049 (0.050) [8/10] & $-$0.029 (0.027) [9/10] & $+$0.38 (1.89) [5/10] & $-$0.002 (0.006) [4/10] & $-$0.312 (0.665) [6/10] \\
        \ours~(joint) $-$ \ours~(video only) & $+$1.57 (1.84) [9/10] & $+$0.063 (0.059) [9/10] & $-$0.025 (0.016) [9/10] & $+$0.68 (0.93) [8/10] & $+$0.001 (0.005) [6/10] & $-$0.010 (0.061) [6/10] \\
        \bottomrule
    \end{tabular}}
\end{table}

\paragraph{Data misfit values for samples in Fig.~\ref{fig:teaser}.} We report the data misfit values in Tab.~\ref{app:teaser-data-misfit}.
\begin{table}[h]
    \centering
    \caption{\textbf{The data misfit values for samples shown in Fig.~\ref{fig:teaser}.} We report the data misfit metrics for the three obtained modes, which achieve comparable data fit, all substantially closer to $1$ than \suppchange{the mean misfit of} any \suppchange{IDM-based} baseline in Tab.~\ref{tab:main-table}.}
    \label{app:teaser-data-misfit}
    \begin{tabular}{l!{\vrule}ccc}
    \toprule
        Metrics & Mode 1 & Mode 2 & Mode 3  \\
         \midrule
       $\bm{\chi}^2_\text{cp}$  & 1.045 & 0.987 & 1.084 \\
       $\bm{\chi}^2_\text{logca}$ & 1.007 & 1.001 &1.202\\
       Data Misfit &  1.026 & 1.007 & 1.143\\
       \bottomrule
    \end{tabular}
\end{table}

\paragraph{\suppchange{Effect of acceleration.}}
\suppchange{Tabs.~\ref{tab:main-table} and~\ref{tab:app-dynamic-mri} report results at 8$\times$ and 6$\times$ acceleration, using the same prior without retraining.}
Compared with the strongest baseline, \ours~improves PSNR by $0.54$ dB at $6\times$ acceleration but by $3.83$ dB at $8\times$ acceleration, \suppchange{suggesting} that the benefit of a spatiotemporal prior grows as the measurements become sparser and the problem more ill-posed.

\begin{figure}
    \centering
    \includegraphics[width=\linewidth]{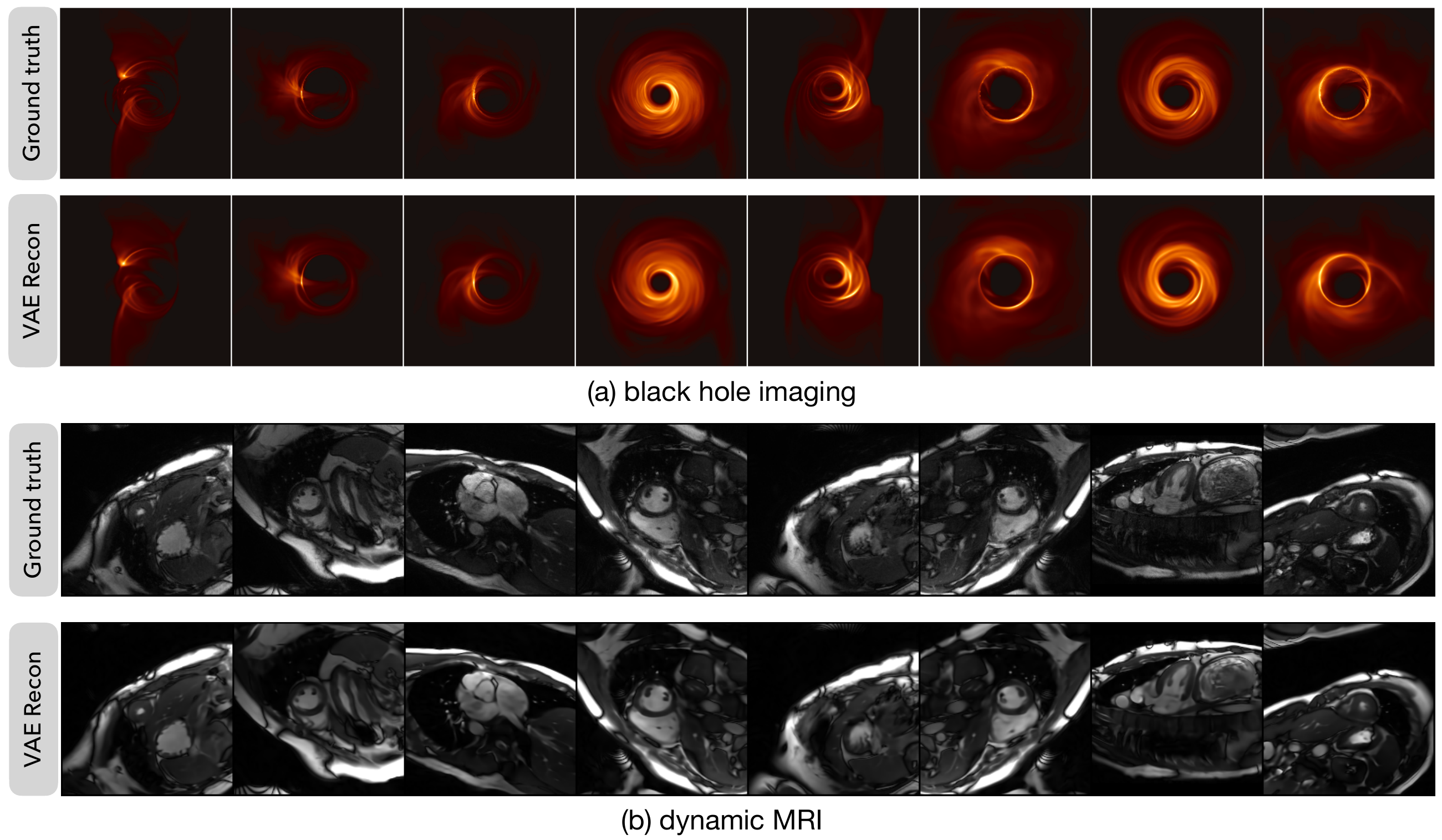}
    \caption{\textbf{Visualization of VAE reconstructions}. The VAE reconstructions are computed by first encoding the ground truth videos and then decoding them.}
    \label{fig:app-vae}
\end{figure}

\begin{figure}
    \centering
    \includegraphics[width=0.9\linewidth]{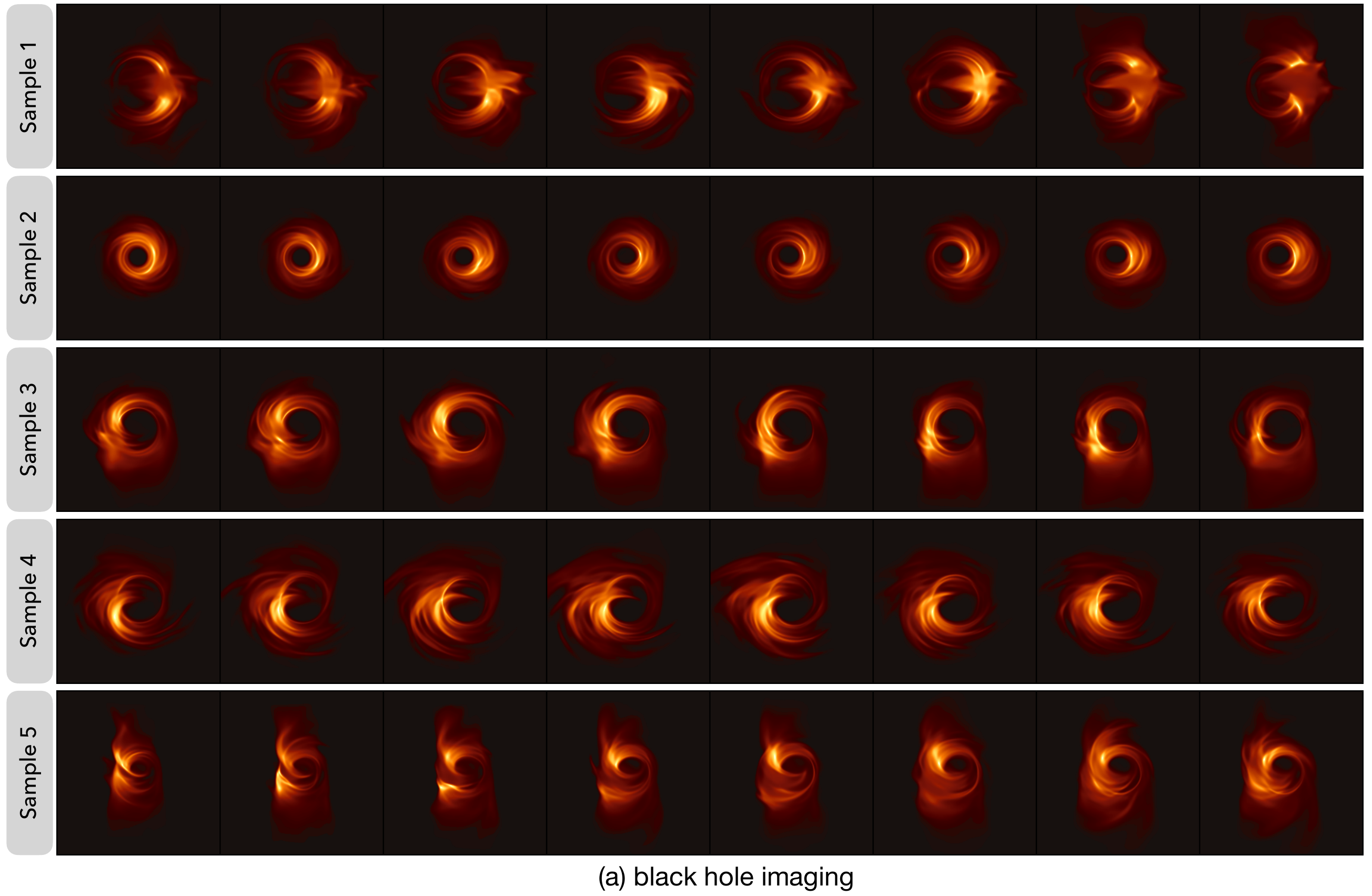}
    \includegraphics[width=0.9\linewidth]{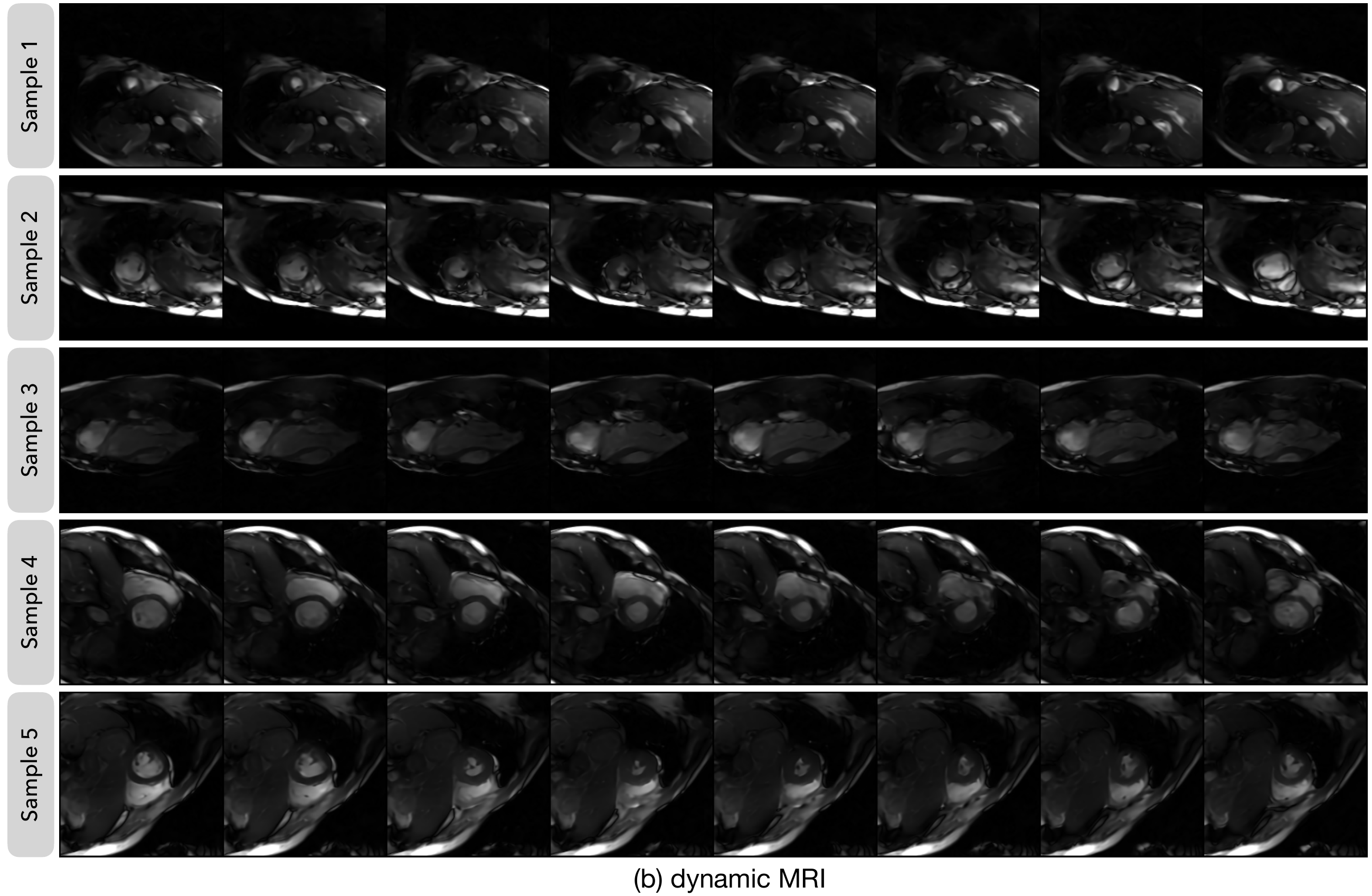}
    \caption{\textbf{Visualization of video diffusion model unconditional samples}. The videos are sampled by solving PF-ODE with 100 Euler's steps.}
    \label{fig:app-prior}
\end{figure}

\begin{figure}
    \centering
    \includegraphics[width=0.8\linewidth]{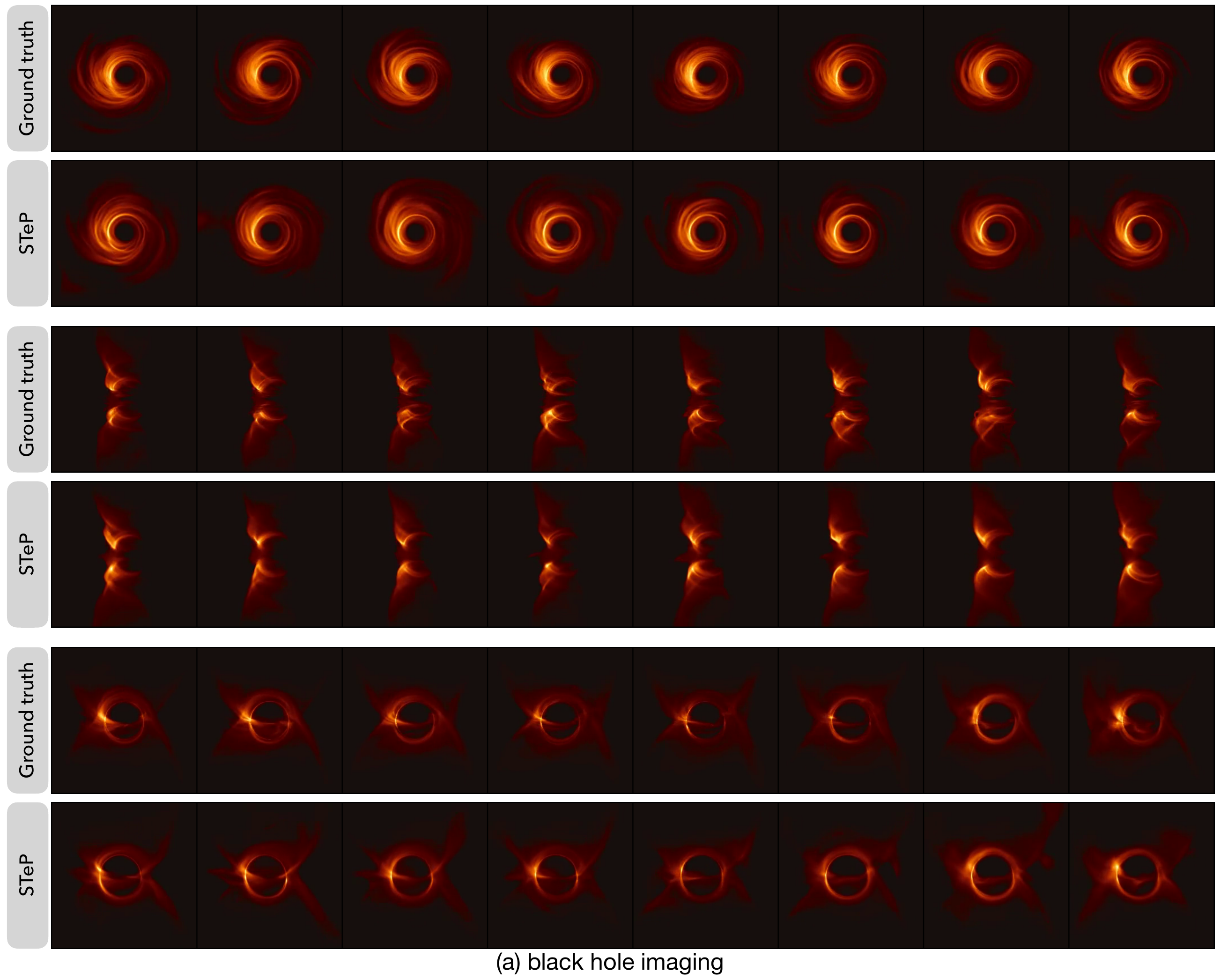}
    \includegraphics[width=0.8\linewidth]{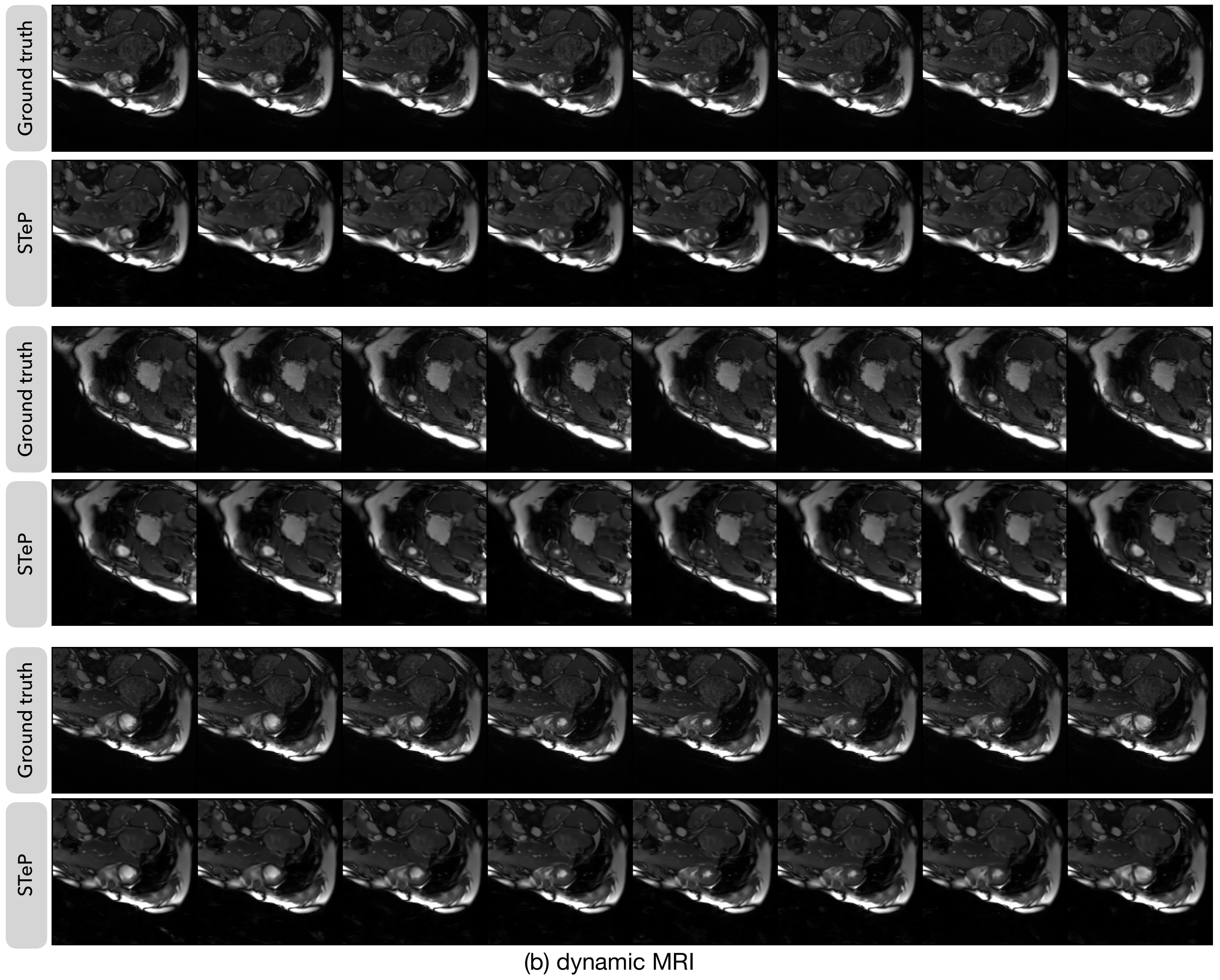}
    \caption{\textbf{Visualization of \ours~posterior samples}. The videos are sampled using \Algref{alg}.}
    \label{fig:app-post}
\end{figure}

\paragraph{More visualizations}
Here, we show the VAE reconstruction results in \Figref{fig:app-vae}, unconditional samples in \Figref{fig:app-prior} and additional posterior samples in \Figref{fig:app-post}.

\end{document}